\documentclass[twoside,11pt]{article}

\usepackage{jmlr2e}
\usepackage{amsmath}
\usepackage{algorithm}
\usepackage{algorithmic}
\usepackage{subfigure}
\usepackage{graphicx}
\usepackage{bbm}
\usepackage{tikz}
\usetikzlibrary{arrows}
\usetikzlibrary{graphs}

\newcommand{\mb}{\mathbf}
\newcommand{\mc}{\mathcal}
\newcommand{\direct}{\textsc{Direct}}
\newcommand{\lipo}{\textsc{LIPO}}
\newcommand{\adalipo}{\textsc{AdaLipo}}
\newcommand{\mcs}{\textsc{MCS}}
\newcommand{\pso}{\textsc{PSO}}

\newcommand{\sce}{\textsc{SCE}}
\newcommand{\de}{\textsc{DE}}
\newcommand{\ga}{\textsc{GA}}
\newcommand{\es}{\textsc{ES}}
\newcommand{\cmaes}{\textsc{CMA-ES}}
\newcommand{\bs}{\boldsymbol}

\ShortHeadings{IFM LAB TUTORIAL SERIES \# 2, COPYRIGHT \copyright IFM LAB}{JIAWEI ZHANG, IFM LAB DIRECTOR}
\firstpageno{1}

\begin{document}

\title{Derivative-Free Global Optimization Algorithms:\\ Bayesian Method and Lipschitzian Approaches}

\author{\name Jiawei Zhang \email jiawei@ifmlab.org \\
	\addr{Founder and Director}\\
       {Information Fusion and Mining Laboratory}\\
       (First Version: March 2019; Revision: April 2019.)}

\maketitle

\begin{abstract}

In this paper, we will provide an introduction to the derivative-free optimization algorithms which can be potentially applied to train deep learning models. Existing deep learning model training is mostly based on the back propagation algorithm, which updates the model variables layers by layers with the gradient descent algorithm or its variants. However, the objective functions of deep learning models to be optimized are usually non-convex and the gradient descent algorithms based on the first-order derivative can get stuck into the local optima very easily. To resolve such a problem, various local or global optimization algorithms have been proposed, which can help improve the training of deep learning models greatly. The representative examples include the Bayesian methods, Shubert-Piyavskii algorithm, {\direct}, {\lipo}, {\mcs}, {\ga}, {\sce}, {\de}, {\pso}, {\es}, {\cmaes}, hill climbing and simulated annealing, etc. One part of these algorithms will be introduced in this paper (including the Bayesian method and Lipschitzian approaches, e.g., Shubert-Piyavskii algorithm, {\direct}, {\lipo} and {\mcs}), and the remaining algorithms (including the population based optimization algorithms, e.g., {\ga}, {\sce}, {\de}, {\pso}, {\es} and {\cmaes}, and random search algorithms, e.g., hill climbing and simulated annealing) will be introduced in the follow-up paper \cite{zhang2019derivativefree_part2} in detail.

\end{abstract}


\tableofcontents

\section{Introduction}\label{sec:intro}

As introduced in the previous IFM Lab tutorial article \cite{zhang2019gradient}, existing deep learning model training is mostly based on the back propagation algorithm, which updates the model variables layers by layers with the gradient descent algorithms. Gradient descent together with its many variants have been shown to be effective for optimizing a large group of problems. However, for the non-convex objective functions of deep learning models, gradient descent algorithms cannot guarantee to identify the globally optimal solutions, whose iterative updating process may inevitably get stuck in the local optima. The performance of gradient descent algorithms are not very robust, which will be greatly degraded for the objective functions with non-smooth shape or learning scenarios polluted by noisy data. Furthermore, the distributed computation process of gradient descent requires heavy synchronization, which may hinder its adoption in large-cluster based distributed computational platforms.

Besides these optimization algorithms based on gradient descent, there also exist a group of optimization algorithms which can be potentially applied to learn the optimal variables for the deep learning models. These optimization algorithms don't need to compute the derivatives of the objective function regarding the model variables, thus they are called the derivative-free optimization algorithms in this paper. The representative examples include the Bayesian methods, Shubert-Piyavskii algorithm, {\direct}, {\lipo}, {\mcs}, {\ga}, {\sce}, {\de}, {\pso}, {\es}, {\cmaes}, hill climbing and simulated annealing, etc. Many of these derivative-free optimization algorithms can even effectively compute the global optimal variables for the non-convex objective functions. These algorithms along can work well for training the deep learning models. Meanwhile, they can also work with the gradient descent algorithms in a hybrid manner similar to Gadam \cite{gadam} as we have introduced in \cite{zhang2019gradient}, which can integrate the advantages of both the gradient descent optimization algorithms together with these derivative-free optimization algorithms.

Here, we will briefly introduce the learning settings as follows. The training set for optimizing the deep learning models can be represented as $\mc{T} = \{(\mb{x}_1, \mb{y}_1), (\mb{x}_2, \mb{y}_2), \cdots, (\mb{x}_n, \mb{y}_n)\}$, which involves $n$ pairs of feature-label instances. Formally, for each data instance, its feature vector $\mb{x}_i \in \mathbb{R}^{d_x}, \forall i \in \{1, 2, \cdots, n\}$ and label vector $\mb{y}_i \in \mathbb{R}^{d_y}, \forall i \in \{1, 2, \cdots, n\}$ are of dimensions $d_x$ and $d_y$ respectively. The deep learning models define a mapping $F(\cdot; \boldsymbol{\theta}): \mc{X} \to \mc{Y}$, which projects the data instances from the feature space $\mc{X}$ to the label space $\mc{Y}$. In the above representation of function $F(\cdot, \bs{\theta})$, vector $\boldsymbol{\theta} \in \Theta$ contains the variables involved in the deep learning model and $\Theta$ denotes the variable inference space. Formally, we can denote the dimension of variable vector $\bs{\theta}$ as $d_{\theta}$, which will be used when introducing the algorithms later. Given one data instance featured by vector $\mb{x}_i \in \mc{X}$, we can denote its prediction label vector by the deep learning model as $\hat{\mb{y}}_i = F(\mb{x}_i; \boldsymbol{\theta})$. Compared against its true label vector $\mb{y}_i$, we can denote the introduced loss for instance $\mb{x}_i$ as $\ell (\hat{\mb{y}}_i, \mb{y}_i)$. Several frequently used loss representations have been introduced in \cite{zhang2019gradient}, and we will not redefine them here again. For all the data instances in the training set, we can represent the total loss term as 
\begin{equation}
\mc{L}(\boldsymbol{\theta}) = \mc{L}(\boldsymbol{\theta}; \mc{T}) = \sum_{(\mb{x}_i, \mb{y}_i) \in \mc{T}} \ell (\hat{\mb{y}}_i, \mb{y}_i).
\end{equation}
And the deep model learning can be formally denoted as the following function:
\begin{equation}
\min_{\boldsymbol{\theta} \in \Theta} \mc{L}(\boldsymbol{\theta}),
\end{equation}
which is also the main objective function to be studied in this paper.

We need to add a remark here, the above objective function defines a minimization problem. Meanwhile, when introducing some of the optimization algorithms in the following sections, we may assume the objective function to be a maximization function instead for simplicity. The above objective can be transformed into a maximization problem easily by introducing a new term $\mc{L}'(\boldsymbol{\theta}) = -\mc{L}(\boldsymbol{\theta})$. We will clearly indicate it when the algorithm is introduced for a maximization problem.

In the following part of this paper, we will introduce the derivative-free optimization algorithms that can be potentially used to resolve the above objective function. To be more specific, this paper covers the introduction to the Bayesian method as well as the Lipschitzian approaches (including Shubert-Piyavskii algorithm, {\direct}, {\lipo} and {\mcs}) for global optimization. The population based algorithms (e.g., {\ga}, {\sce}, {\de}, {\pso}, {\es} and {\cmaes}) and the random search based optimization algorithms (e.g., hill climbing and simulated annealing) will be introduced in the follow-up article \cite{zhang2019derivativefree_part2} in detail.


\section{Bayesian Method for Global Optimization}

Bayesian methods \cite{bayesian_methods} assume the objective function is a black-box. Evaluating the objective function is expensive or even impossible, whose derivatives and convexity properties are also unknown. Such an assumption holds for some deep learning models involving complicated structures with multiple layers of non-linear projections, since its analytical expression and derivatives will be too complex to analyze. Instead of optimizing the objective function directly, Bayesian methods propose to approximate the objective function based on a set of sampled points in the variable domain, and tries to select the optimal variable based on the approximated objective function. Such a process involves two basic functions in the Bayesian methods: \textit{surrogate function} and \textit{acquisition function}, respectively. To approximate the objective function, Bayesian method adopts a \textit{surrogate function} to update the posterior distribution of the function based on both the prior distribution and the likelihood computed with the sampled points. Meanwhile, to sample efficiently, Bayesian methods use an \textit{acquisition function} to determine the next optimal variable to sample from the variable domain.

In Algorithm~\ref{alg:chapter_derivative_free_bayesian_method}, we provide the pseudo-code of the Bayesian Method, which involves 3 main steps: (1) optimal variable point computing via maximizing the acquisition function; (2) objective function value sampling subject to noise; and (3) Gaussian Process updating with the surrogate function to update $\mu(\boldsymbol{\theta})$ and $\sigma^2(\boldsymbol{\theta})$.

\begin{algorithm}[H]
\small
\caption{Bayesian Optimization Methods}
\label{alg:chapter_derivative_free_bayesian_method}
\begin{algorithmic}[1]
	\REQUIRE Input variable space $\Theta$; GP prior $\mu$, $\sigma$.
\ENSURE  Model Parameter $\boldsymbol{\theta}$
\FOR	{$\tau = 1, 2, \cdots$}
\STATE	{Find $\boldsymbol{\theta}_{\tau} = \arg \max_{\boldsymbol{\theta} \in \Theta} \alpha(\boldsymbol{\theta}; \mc{S}_{\tau-1})$ by optimizing the acquisition function.}
\STATE	{Sample the objective function value $z_{\tau} = \mc{L}(\boldsymbol{\theta}_{\tau}) + \epsilon_{\tau}$.}
\STATE	{Update $\mc{S}_{\tau} = \mc{S}_{\tau-1} \cup \{(\boldsymbol{\theta}_{\tau}, z_{\tau})\}$ and recompute $\mu(\boldsymbol{\theta})$, $\sigma^2(\boldsymbol{\theta})$ based on the surrogate function.}
\ENDFOR
\STATE	{\textbf{Return} model variable $\boldsymbol{\theta}$}
\end{algorithmic}
\end{algorithm}

Most of the steps in Algorithm~\ref{alg:chapter_derivative_free_bayesian_method} will be covered in the following subsections, except for the noisy function value sampling part. Considering noise-free observations are very hard to achieve in the real-world. In the algorithm, term $\epsilon_\tau \sim \mc{N}(0, \sigma^2_{\textsc{noise}})$ is a random noise sampled from the normal distribution, which performs a noisy transformation on function $\mc{L}(\boldsymbol{\theta})$. A tutorial on Bayesian methods is available at \cite{bayesian_methods_tutorial}.


\subsection{Surrogate Function for Posterior Distribution Updating}

In Bayesian methods, the sample sequence achieved prior to iteration $\tau \ge 1$ can be denoted as $\mc{S}_{\tau-1} = \{\boldsymbol{\theta}_1, \boldsymbol{\theta}_2, \cdots, \boldsymbol{\theta}_{\tau-1}\}$ (or $\mc{S}_{\tau-1} = \{(\boldsymbol{\theta}_1, z_1), (\boldsymbol{\theta}_2, z_2), \cdots, (\boldsymbol{\theta}_{\tau-1}, z_{\tau -1})\}$ if we pair the sampled variables with their introduced loss terms together), where $z_i = \mc{L}(\boldsymbol{\theta}_i)$ denotes the loss value introduced by variable $\boldsymbol{\theta}_i$. Formally, let $P(\mc{L})$ denote the prior distribution of the loss function, and $P(\mc{S}_{\tau-1} | \mc{L})$ denote the likelihood function of achieve the sampled variables based on the loss function $\mc{L}$. Based on these two terms, the posterior distribution of the objective loss function based on the sampled points can be denoted as
\begin{equation}
P(\mc{L}|\mc{S}_{\tau-1}) \propto P(\mc{S}_{\tau-1} | \mc{L}) P(\mc{L}).
\end{equation}
The step of updating the posterior distribution of the loss function based on the sampled variable points is also interpreted as estimating the objective function with a \textit{surrogate function} (also called a \textit{response surface}). 


As pointed out in \cite{bayesian_methods}, such a process can be modeled effectively with the Gaussian Process (GP), where the prior distribution of the loss function is subject to a Gaussion distribution. Here, let's misuse the loss function notation $\mc{L}$ as a variable, and it follows
\begin{equation}
\mc{L}(\boldsymbol{\theta}) \sim \mc{N}(m_{\boldsymbol{\theta}}, k_{\boldsymbol{\theta}}),
\end{equation}
where $m_{\boldsymbol{\theta}}$ and $k_{\boldsymbol{\theta}}$ denote the mean and co-variance functions of the normal distribution respectively. For convenience, the prior mean is usually assumed to be zeros, and the variance function $k_{\boldsymbol{\theta}}$ has several different choices (we will introduce later).

Following such a distribution, let's assume we have sampled $\tau-1$ variable points already from objective variable domain $\Theta$, whose loss terms can be denoted as a vector $[\mc{L}(\boldsymbol{\theta}_1), \mc{L}(\boldsymbol{\theta}_2), \cdots, \mc{L}(\boldsymbol{\theta}_{\tau-1})]^\top$. According to the above descriptions, they should follow the following multivariate normal distribution
\begin{equation}
\begin{bmatrix}
\mc{L}(\boldsymbol{\theta}_1)\\
\mc{L}(\boldsymbol{\theta}_2)\\
\vdots\\
\mc{L}(\boldsymbol{\theta}_{\tau-1})
\end{bmatrix}
\sim
\mc{N}\left(\mb{0}, \mb{K}\right),
\end{equation}
where the covariance matrix 
\begin{equation}
\mb{K} =
\begin{bmatrix}
k(\boldsymbol{\theta}_1, \boldsymbol{\theta}_1), & \cdots, & k(\boldsymbol{\theta}_1, \boldsymbol{\theta}_{\tau-1})\\
\vdots & \ddots & \vdots \\
k(\boldsymbol{\theta}_{\tau-1}, \boldsymbol{\theta}_1), & \cdots, &k(\boldsymbol{\theta}_{\tau-1}, \boldsymbol{\theta}_{\tau-1})
\end{bmatrix}
\end{equation}

By further sampling one more variable point $\bs{\theta}_{\tau}$, in a similar way, we can achieve the distribution of the sampled $\tau$ variable points as 
\begin{equation}
\begin{bmatrix}
\mc{L}(\boldsymbol{\theta}_1)\\
\mc{L}(\boldsymbol{\theta}_2)\\
\vdots\\
\mc{L}(\boldsymbol{\theta}_{\tau-1})\\
\mc{L}(\boldsymbol{\theta}_{\tau})
\end{bmatrix}
\sim
\mc{N}\left(\mb{0}, 
\begin{bmatrix}
\mb{K} & \mb{k}\\
\mb{k}^\top & k(\boldsymbol{\theta}_{\tau}, \boldsymbol{\theta}_{\tau})
\end{bmatrix}
\right),
\end{equation}
where vector $\mb{k} = [k(\boldsymbol{\theta}_{1}, \boldsymbol{\theta}_{\tau}), k(\boldsymbol{\theta}_{2}, \boldsymbol{\theta}_{\tau}), \cdots, k(\boldsymbol{\theta}_{\tau-1}, \boldsymbol{\theta}_{\tau})]^\top$.

Based on the Sherman-Morrison-Woodbury formula \cite{sherman_morrison_woodbury_formula}, we can represent the posterior distribution of the loss function for the sampled point $\boldsymbol{\theta}_{\tau}$ as follows
\begin{equation}
P(\mc{L}(\boldsymbol{\theta}_{\tau}) | \mc{S}_{\tau-1}, \boldsymbol{\theta}_{\tau}) \sim \mc{N}(\mu_{\tau}(\boldsymbol{\theta}_{\tau}), \sigma^2_{\tau}(\boldsymbol{\theta}_{\tau})),
\end{equation}
where 
\begin{align}\label{equ:bayesian_method_subscript}
\mu_{\tau}(\boldsymbol{\theta}_{\tau}) &= \mb{k}^\top \mb{K}^{-1} \begin{bmatrix} \mc{L}(\boldsymbol{\theta}_1)\\
\mc{L}(\boldsymbol{\theta}_2)\\
\vdots\\
\mc{L}(\boldsymbol{\theta}_{\tau-1}) \end{bmatrix},\\
\sigma^2_{\tau}(\boldsymbol{\theta}_{\tau}) &= k(\boldsymbol{\theta}_{\tau}, \boldsymbol{\theta}_{\tau}) - \mb{k}^\top \mb{K}^{-1}\mb{k}.
\end{align}

As we mentioned before, the co-variance function $k(\cdot, \cdot)$ may be defined in different ways, and some representative examples include:
\begin{itemize}
\item \textit{squared exponential function}:
\begin{equation}
k(\boldsymbol{\theta}_i, \boldsymbol{\theta}_j) = \exp(-\frac{1}{2} \left\| \boldsymbol{\theta}_i - \boldsymbol{\theta}_j \right\|^2).
\end{equation}

\item \textit{squared exponential function with hyperparameters}:
\begin{equation}
k(\boldsymbol{\theta}_i, \boldsymbol{\theta}_j) = \exp(-\frac{1}{2\beta^2} \left\| \boldsymbol{\theta}_i - \boldsymbol{\theta}_j \right\|^2),
\end{equation}
where the hyperparameter $\beta$ controls the width of the variance.

\item \textit{squared exponential function with automatic relevance determination hyperparameters}:
\begin{equation}
k(\boldsymbol{\theta}_i, \boldsymbol{\theta}_j) = \exp(-\frac{1}{2} (\boldsymbol{\theta}_i - \boldsymbol{\theta}_j)^\top diag(\boldsymbol{\beta})^{-2} (\boldsymbol{\theta}_i - \boldsymbol{\theta}_j) ),
\end{equation}
where $\boldsymbol{\beta}$ is a hyperparameter vector and notation $diag(\boldsymbol{\beta})$ denotes a diagonal matrix with entries $\boldsymbol{\beta}$ on its diagonal.

\item \textit{Mat\'ern function}:
\begin{equation}
k(\boldsymbol{\theta}_i, \boldsymbol{\theta}_j) = \frac{1}{2^{\varsigma -1}\Gamma(\varsigma)} (2\sqrt{\varsigma} \left\| \boldsymbol{\theta}_i- \boldsymbol{\theta}_j \right\|)^{\varsigma} H_{\varsigma} (2\sqrt{\varsigma} \left\| \boldsymbol{\theta}_i- \boldsymbol{\theta}_j  \right\|),
\end{equation}
where $\Gamma(\cdot)$ and $H_{\varsigma}(\cdot)$ are the Gamma function and Bessel function of order $\varsigma$ respectively. As the order hyperparameter $\varsigma \to \infty$, the \textit{Mat\'ern function} will be reduced to the \textit{squared exponential function}.
\end{itemize}


\subsection{Acquisition Function for Optimal Variable Sampling}

Based on the posterior distribution updated via the Gaussian Process iteratively, here, we will introduce the \textit{acquisition function}, which guides the search for the optimal variables in Bayesian methods. Typically, acquisition functions are defined to ensure high acquisition function values correspond to potentially higher values of the objective function. Therefore, the optimal variable points will be selected iteratively which can maximize the acquisition function. Considering the objective function studied in deep learning model training is a minimization problem, which can be transformed into a maximization problem easily by defining a pseudo-loss function $\mc{L}'(\boldsymbol{\theta}) = -\mc{L}(\boldsymbol{\theta})$. The following contents are introduced based on the maximization objective function on the loss function $\mc{L}'(\boldsymbol{\theta})$ (to simplify the notations, we will use $\mc{L}(\boldsymbol{\theta})$ directly to denote the function to be maximized).

Formally, based on the sampling records prior to iteration $\tau$, i.e., $\mc{S}_{\tau-1} = \{\boldsymbol{\theta}_1, \boldsymbol{\theta}_2, \cdots, \boldsymbol{\theta}_{\tau-1}\}$, Bayesian methods will select the optimal variable for iteration $\tau$ with the following function:
\begin{equation}
\boldsymbol{\theta}_{\tau} = \arg \max_{\boldsymbol{\theta} \in \Theta} \alpha(\boldsymbol{\theta} ; \mc{S}_{\tau - 1}),
\end{equation}
where $\alpha(\boldsymbol{\theta} ; \mc{S}_{\tau - 1})$ is called the \textit{acquisition function}. {Acquisition function} helps the Bayesian methods to compute the next sampling point of the variables, and it has various representations. Popular acquisition functions include the \textit{maximum probability of improvement} (MPI) \cite{mpi}, \textit{expected improvement} (EI) \cite{ei} and \textit{upper confidence bound} (UCB) \cite{ucb}, which will be introduced as follows respectively. Meanwhile, the above objective acquisition function can be optimized with the {\direct} algorithm as introduced in Section~\ref{subsec:direct}.


\subsubsection{MPI}

Formally, let $\boldsymbol{\theta}^+ = \arg\max_{\boldsymbol{\theta} \in \mc{S}_{\tau -1}} \mc{L}(\boldsymbol{\theta})$ denote the best variable among all the variables we have sampled in the previous $\tau-1$ iterations. For the MPI acquisition function \cite{mpi}, its main objective will be to select the variable points which can lead to the improvement with the maximum probability. In other words, the MPI based acquisition function can be formally represented as
\begin{align}
\alpha_{\textsc{mpi}}(\boldsymbol{\theta}) &= P(\mc{L}\left(\boldsymbol{\theta}) \ge \mc{L}(\boldsymbol{\theta}^+)\right)\\
&= \Phi \left( \frac{\mu(\boldsymbol{\theta}) - \mc{L}(\boldsymbol{\theta}^+)}{\sigma(\boldsymbol{\theta})} \right),
\end{align}
where function $\Phi(\cdot)$ denote the cumulative distribution function (CDF) of the standard Gaussian distribution and the iteration index subscript (i.e., $\tau$ as we used in Equations~\ref{equ:bayesian_method_subscript}) on the mean and standard deviation functions $\mu(\boldsymbol{\theta})$ and $\sigma(\boldsymbol{\theta})$ are omitted for simplicity reasons.

According to the above function, the MPI based acquisition function mainly care about the probability value instead of the real advantages of the loss term $\mc{L}(\boldsymbol{\theta})$ against $\mc{L}(\boldsymbol{\theta}^+)$. It is possible that maximizing the above acquisition function may lead to a variable point $\boldsymbol{\theta}$ which is slightly greater than $\boldsymbol{\theta}^+$ (i.e., $\mc{L}(\boldsymbol{\theta}) - \mc{L}(\boldsymbol{\theta}^+ \to 0^+$) simply because the probability value $P(\mc{L}\left(\boldsymbol{\theta}) \ge \mc{L}(\boldsymbol{\theta}^+)\right)$ is large. To resolve such a problem, some research works also propose to use the following MPI based acquisition function instead:
\begin{align}
\alpha_{\textsc{mpi}}(\boldsymbol{\theta}) &= P(\mc{L}\left(\boldsymbol{\theta}) \ge \mc{L}(\boldsymbol{\theta}^+)+ \xi\right)\\
&= \Phi \left( \frac{\mu(\boldsymbol{\theta}) - \mc{L}(\boldsymbol{\theta}^+) - \xi}{\sigma(\boldsymbol{\theta})} \right).
\end{align}
The parameter $\xi$ defines the desired gap in the loss function in the variable selection, which is usually a hyperparameter inputted into the algorithm.


\subsubsection{EI}

Since $\mc{L}(\boldsymbol{\theta})$ is the objective function to optimize in our studied problem, the main focus in variable selection should be picking the ones which can lead the maximum improvement against the best historical variable, i.e., $\boldsymbol{\theta}^+ = \arg\max_{\boldsymbol{\theta} \in \mc{S}_{\tau -1}} \mc{L}(\boldsymbol{\theta})$. In other words, the EI \cite{ei} based acquisition function can be defined as follows:
\begin{align}
\alpha_{\textsc{ei}}(\boldsymbol{\theta}) &= \mathbb{E}\left( I(\boldsymbol{\theta}; \mc{S}_{\tau-1}) \right)\\
&=  \mathbb{E}\left( \max\{ 0, \mc{L}(\boldsymbol{\theta}) - \mc{L}(\boldsymbol{\theta}^+) \} \right),
\end{align}
where $I(\mc{L}(\boldsymbol{\theta}); \mc{S}_{\tau-1}) = \max\{0, \mc{L}(\boldsymbol{\theta}) - \mc{L}(\boldsymbol{\theta}^+)\}$ is also called the \textit{improvement function}. 

Meanwhile, the normal density function of $I$ (here, we treat function $I(\boldsymbol{\theta}; \mc{S}_{\tau-1})$ as a variable) can be computed based on the posterior distribution parameterized by mean $\mu(\boldsymbol{\theta})$ and co-variance $\sigma^2(\boldsymbol{\theta})$ as follows:
\begin{equation}
P(I) = \frac{1}{\sqrt{2\pi} \sigma(\boldsymbol{\theta})} \exp\left( - \frac{ \left(\mu(\boldsymbol{\theta}) - \mc{L}(\boldsymbol{\theta}^+) - I \right)^2 }{2 \sigma^2(\boldsymbol{\theta})} \right).
\end{equation}
Therefore, the expected improvement can be represented as
\begin{align}
\mathbb{E}(I) &= \int_{I=0}^{I = \infty} I P(I) \mathrm{d} I\\
&= \sigma(\boldsymbol{\theta}) \left[ \frac{\mu(\boldsymbol{\theta}) - \mc{L}(\boldsymbol{\theta}^+)}{\sigma(\boldsymbol{\theta})} \cdot \Phi\left( \frac{\mu(\boldsymbol{\theta}) - \mc{L}(\boldsymbol{\theta}^+)}{\sigma(\boldsymbol{\theta})}  \right) + \phi\left( \frac{\mu(\boldsymbol{\theta}) - \mc{L}(\boldsymbol{\theta}^+)}{\sigma(\boldsymbol{\theta})}  \right) \right],
\end{align}
where $\Phi(\cdot)$ and $\phi(\cdot)$ denote the CDF and PDF of the standard Gaussian distribution respectively. 

Based on the above analysis, we can formally represent the EI based acquisition function as follows:
\begin{equation}
\alpha_{\textsc{ei}}(\boldsymbol{\theta}) = 
\begin{cases}
\left(\mu(\boldsymbol{\theta}) - \mc{L}(\boldsymbol{\theta}^+) \right) \Phi(Z) + \sigma(\boldsymbol{\theta}) \phi(Z) & \mbox{ if }\sigma(\boldsymbol{\theta}) > 0;\\
0 & \mbox{ if }\sigma(\boldsymbol{\theta}) = 0,
\end{cases}
\end{equation}
where term $Z= \frac{\mu(\boldsymbol{\theta}) - \mc{L}(\boldsymbol{\theta}^+)}{\sigma(\boldsymbol{\theta})}$.


\subsubsection{UCB}

As introduced in \cite{ucb}, given a random function on variable $\boldsymbol{\theta}$, we can compute its maximum value by maximizing its \textit{upper confidence bound} (UCB) as follows:
\begin{equation}
UCB(\boldsymbol{\theta}) = \mu(\boldsymbol{\theta}) + \kappa \sigma(\boldsymbol{\theta}),
\end{equation}
where $\kappa \ge 0$ is a hyperparameter.

In recent years, some works propose to adopt the UCB function to define the acquisition function for optimizing the posterior distribution function. Formally, let $\mc{L}(\boldsymbol{\theta}^*)$ denote the globally optimal value. Therefore, for any variable $\boldsymbol{\theta}$ we choose, we can represent its distance compared with the optimal value as the following regret function \cite{ucb_regret}:
\begin{equation}
r(\boldsymbol{\theta}) = \mc{L}(\boldsymbol{\theta}^*) - \mc{L}(\boldsymbol{\theta}).
\end{equation}

In sampling the variable points, our main objective will be to minimize the regret function, which is equivalent to the maximization of function values at the sampled variable points, i.e.,
\begin{equation}
\min \sum_{i = 1}^{i = \tau} r(\boldsymbol{\theta}_i) \propto \max \sum_{i = 1}^{i = \tau} \mc{L}(\boldsymbol{\theta}_i).
\end{equation}

By using the \textit{upper confidence bound} as the selection criterion, we can define the UCB based acquisition function as follows:
\begin{equation}
\alpha_{\textsc{ucb}}(\boldsymbol{\theta}) = \mu(\boldsymbol{\theta}) + \sqrt{\nu \gamma_{\tau}} \sigma(\boldsymbol{\theta}),
\end{equation}
where $\gamma_{\tau} = 2 \log(d_{\theta} \tau^2 \pi^2 /6\delta)$ is a function about $\tau$, while $\delta \in (0,1)$ and $\nu > 0$ are the hyperparameters in the algorithm. By maximizing the UCB based acquisition function, we will be able to select promising variable points iteratively.


\section{Lipschitzian Approaches for Global Optimization}

Lipschitzian approach is the name for a group of optimization approaches based on the Lipschitz functions. Lipschitzian approaches usually require the prior knowledge about the Lipschitz constant, which is a bound on the change rate of the objective function. Meanwhile, in the case where the Lipschitz constant is unknown, some other variant approaches have been proposed, like {\direct}, {\lipo} and {\mcs}, which will be introduced in this section as well. Lipschitzian approaches have many outstanding advantages compared against other optimization algorithms. First of all, Lipschitzian approaches have very few parameters to tune, which is a desired strength compared with other algorithms. On the other hand, since they are the deterministic methods, Lipschitzian approaches don't require multiple runs in computing the optimum. 


\subsection{Shubert-Piyavskii Algorithm}

In this part, we will introduce the classic Shubert-Piyavskii algorithm \cite{shubert, piyavskii} based on an objective function with one single variable. For the case where the objective function is multivariate, several extension works have been proposed, and we will briefly mention some of them at the end of this subsection. Formally, let's assume the objective function to be minimized here is $\mc{L}(\theta)$, where the variable $\theta$ is within a closed interval $[a, b]$ (i.e., variable $\theta$ is constrained and $a$, $b$ denote the upper bound and lower bound respectively).

Standard Lipschitzian approaches usually assume the objective function $\mc{L}(\theta)$ is subject to the Lipschitz function, i.e., there exist a finite bound on the rate of function changes. Formally, assuming there exist a positive constant $K$, the following Lipschitz function should hold:
\begin{equation}\label{equ:lipschitz_function}
\left| \mc{L}(\theta_1) - \mc{L}(\theta_2) \right| \le K \cdot \left| \theta_1 - \theta_2 \right|, \forall \theta_1, \theta_2 \in [a, b],
\end{equation}
where constant $K$ is also called the Lipschitz constant.

The above inequality actually creates a lower bound for the objective function, and we will illustrate it with the following example.
\begin{example}

\begin{figure}[t]
    \centering
    \includegraphics[width=0.6\textwidth]{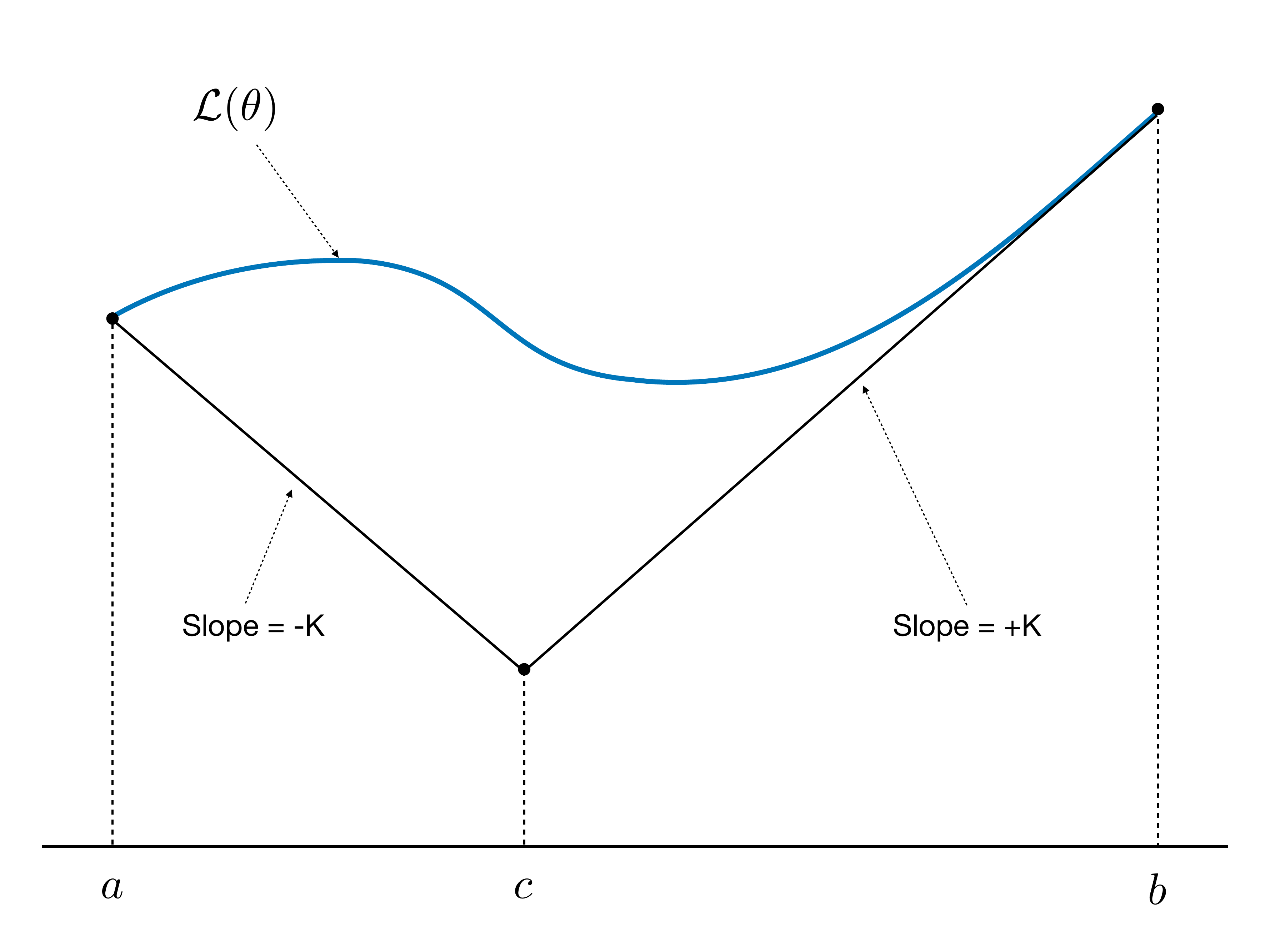}
    \caption{An Example to Illustrate the Shubert-Piyavskii Algorithm.}
    \label{fig:chap_derivative_free_shubert}
\end{figure}

Based on the Lipschitz function provided in formula~(\ref{equ:lipschitz_function}), we can replace $\theta_1$ with $\theta$ and $\theta_2$ with $a$, which will lead to
\begin{align}\label{equ:lipschitz_function_left_endpoint}
&\left| \mc{L}(\theta) - \mc{L}(a) \right| \le K \cdot \left| \theta - a \right|\\
&\Rightarrow \mc{L}(a) - \mc{L}(\theta) \le \left| \mc{L}(\theta) - \mc{L}(a) \right| \le K \cdot (\theta - a)\\
&\Rightarrow \mc{L}(\theta) \ge \mc{L}(a) - K \cdot (\theta - a), \forall \theta \in [a, b].
\end{align} 

Similarly, by substituting $\theta_1$ with $\theta$ and $\theta_2$ with $b$, we can get
\begin{equation}\label{equ:lipschitz_function_right_endpoint}
\mc{L}(\theta) \ge \mc{L}(b) + K \cdot (\theta - b), \forall \theta \in [a, b].
\end{equation}

These two inequalities introduce two lines with slopes $-K$ and $+K$ as shown in Figure~\ref{fig:chap_derivative_free_shubert} respectively, which forms a ``V'' structure. According to the plot, the curve of function $\mc{L}(\theta)$ is above these two lines, i.e., function $\mc{L}(\theta)$ is lower-bounded by these two lines. The intersection of these two lines is the lowest point for these two lines, whose coordinate can be denoted as $c$. With some simple derivations, we can get the representations of $c$ as well as its corresponding function value to be
\begin{align}\label{equ:shubert-piyavskii-division-point}
c &= \frac{(a + b)}{2} + \frac{\mc{L}(a) - \mc{L}(b)}{2K},\\
\mc{L}(c) &= \frac{\mc{L}(a) + \mc{L}(b)}{2} - K \cdot (b - a).
\end{align}
\end{example}

\begin{figure}[t]
\begin{minipage}[b]{\linewidth}
\centering
\subfigure[Step 1]{\label{fig:chap_derivative_free_shubert_process_step_a}
    \begin{minipage}[l]{0.4\columnwidth}
      \centering
      \includegraphics[width=1.0\textwidth]{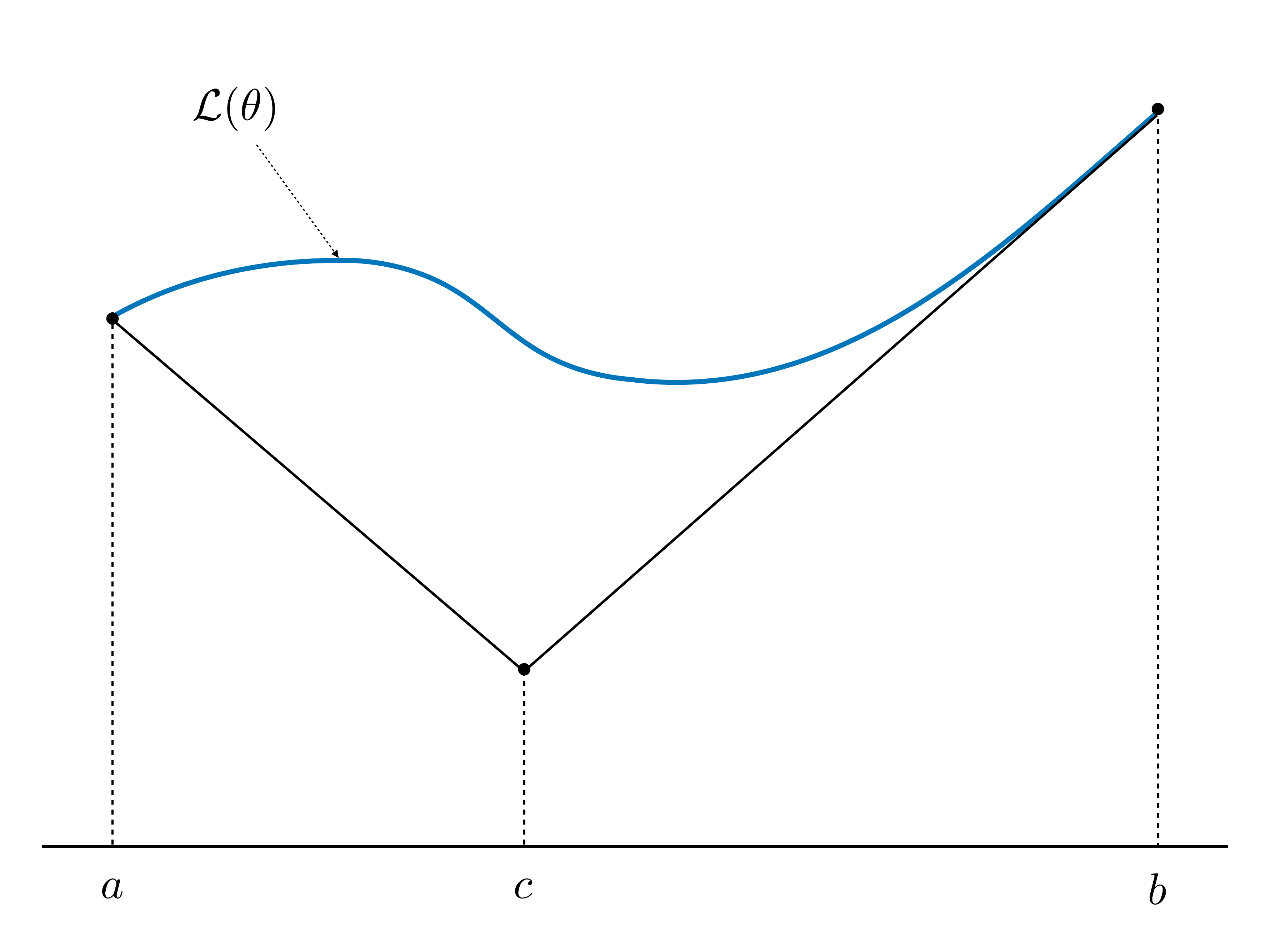}
    \end{minipage}
}
\subfigure[Step 2]{ \label{fig:chap_derivative_free_shubert_process_step_b}
    \begin{minipage}[l]{0.4\columnwidth}
      \centering
      \includegraphics[width=1.0\textwidth]{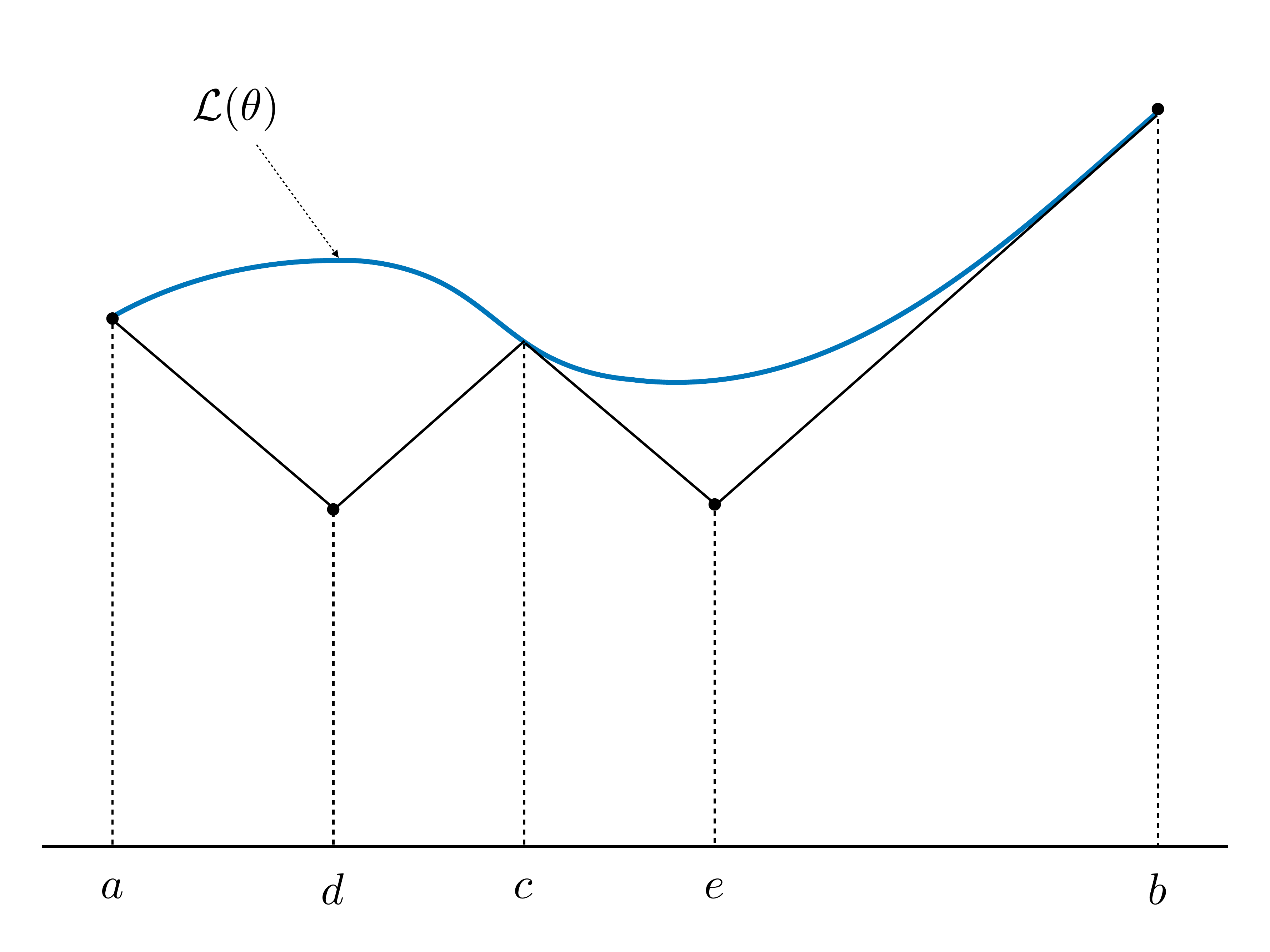}
    \end{minipage}
}
\subfigure[Step 3]{ \label{fig:chap_derivative_free_shubert_process_step_c}
    \begin{minipage}[l]{0.4\columnwidth}
      \centering
      \includegraphics[width=1.0\textwidth]{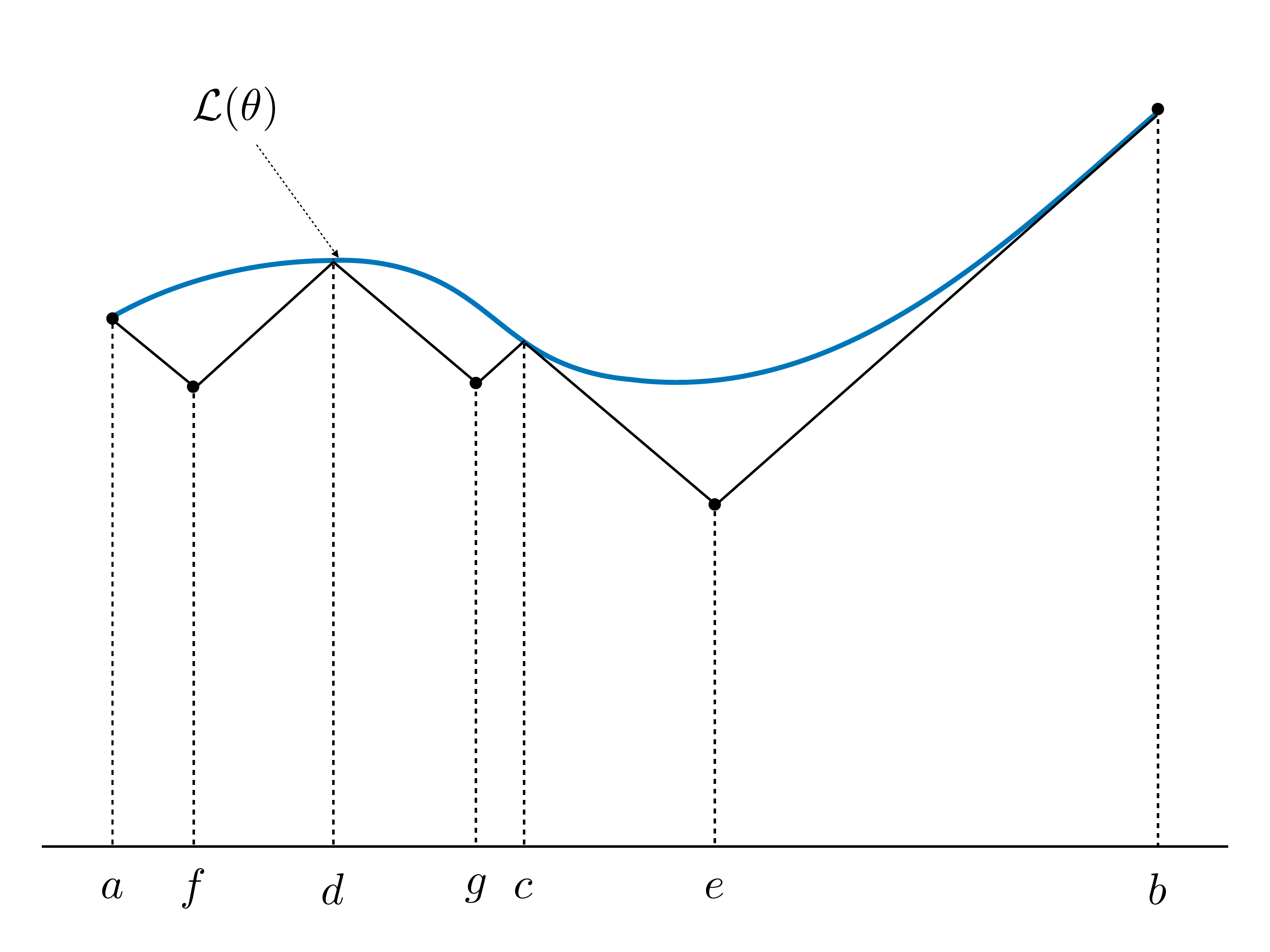}
    \end{minipage}
}
\caption{Shubert-Piyavskii Algorithm Optimization Process.}\label{fig:chap_derivative_free_shubert_process}
\end{minipage}
\end{figure}

The above two equations will serve as the core component in the Shubert-Piyavskii algorithm, whose detailed descriptions will be provided as follows. We can still use the curve in Figure~\ref{fig:chap_derivative_free_shubert} as an example. 
\begin{enumerate}
\item As shown by Figure~\ref{fig:chap_derivative_free_shubert_process_step_a}, the algorithm starts by the evaluation of the objective function at the endpoints $a$ and $b$ to compute terms $\mc{L}(a)$ and $\mc{L}(b)$, which help determine the division point $c$ together with its evaluation term $\mc{L}(c)$. 
\item Point $c$ divides the search space $[a, b]$ into two intervals $[a, c]$ and $[c, b]$, and we are interested in which interval may have a lower objective function value. As shown in Figure~\ref{fig:chap_derivative_free_shubert_process_step_b}, the evaluation at the endpoints of them will bring about two new division points $d$ and $e$ together with their evaluations $\mc{L}(d)$ and $\mc{L}(e)$ respectively. In this example, we happen to have $\mc{L}(d)=\mc{L}(e)$, and the algorithm will randomly choose one interval to divide, e.g., interval $[a, c]$, which will lead to $3$ intervals $[a, d]$, $[d, c]$ and $[c, b]$ respectively. 
\item Next, as shown in Figure~\ref{fig:chap_derivative_free_shubert_process_step_c}, the algorithm will compare the evaluations at the division points of these three intervals, i.e., $f$, $g$ and $e$, and chose the next ones to explore. Among these three intervals, $[c, b]$ will lead to a smaller objective function value at the division point $e$ and should be explored next.
\end{enumerate}
Such an algorithm continues until the minimum of the approximation is within some pre-specified tolerance of the current best solution. The pseudo-code of the Shubert-Piyavskii algorithm is provided in Algorithm~\ref{alg:shubert_piyavskii_algorithm}.

In the algorithm, the interval selection criteria is the objective function evaluation at the corresponding division point, as illustrated in Equation~\ref{equ:shubert-piyavskii-division-point}. Among all the existing intervals, the one with the minimal $\mc{L}(c) = \frac{\mc{L}(a) + \mc{L}(b)}{2} - K \cdot (b - a)$ will be picked. By checking the representation of $\mc{L}(c)$, we observe that $\mc{L}(c)$ can achieve a smaller value iff (1) $\frac{\mc{L}(a) + \mc{L}(b)}{2}$ is small, and (2) $K \cdot (b - a)$ is large. In other words, for the intervals whose endpoints correspond to smaller objective function values (i.e., the interval contains points with relatively smaller values), and the interval is wide (i.e., there exists more space to explore in the interval), they will be picked for exploration in the Shubert-Piyavskii algorithm. 

\begin{algorithm}[t]
\small
\caption{Shubert Piyavskii Algorithm}
\label{alg:shubert_piyavskii_algorithm}
\begin{algorithmic}[1]
	\REQUIRE Variable range [a, b].
\ENSURE  Model Parameter ${\theta}$
\STATE	{Evaluate $\mc{L}(a)$ and $\mc{L}(b)$ for endpoints $a$ and $b$.}
\STATE	{Compute division point $c$ and evaluate $\mc{L}(c)$ according to Equation~(\ref{equ:shubert-piyavskii-division-point}).}
\STATE	{Initialize set $\mc{S} = \{[\mc{L}(c): (a, b; c)]\}$.}
\WHILE	{Stop criteria is not met.}
\STATE	{Select interval $[a_i, b_i]$ whose corresponding $\mc{L}(c_i)$ is the minimum.}
\STATE	{$\mc{S} = \mc{S} \setminus \{[\mc{L}(c_i): (a_i, b_i; c_i)]\}$.}
\STATE	{Divide interval $[a_i, b_i]$ into two intervals $[a_i, c_i]$ and $[c_i, b_i]$.}
\STATE	{Compute division point $d_i$ and evaluate $\mc{L}(d_i)$ for interval $[a_i, c_i]$ according to Equation~(\ref{equ:shubert-piyavskii-division-point}).}
\STATE	{Compute division point $e_i$ and evaluate $\mc{L}(e_i)$ for interval $[c_i, b_i]$ according to Equation~(\ref{equ:shubert-piyavskii-division-point}).}
\STATE	{Update $\mc{S} = \mc{S} \cup \{[\mc{L}(d_i): (a_i, c_i; d_i)], [\mc{L}(e_i): (c_i, b_i; e_i)]\}$.}
\ENDWHILE
\end{algorithmic}
\end{algorithm}

The Shubert-Piyavskii algorithm has two main disadvantages in applications in the real world: 

\begin{itemize}

\item \textbf{Global Exploration}: The Shubert-Piyavskii algorithm is highly dependent on the hyperparameter $K$. To ensure the Lipschitz function can hold, the parameter $K$ is usually a very large value. It also brings about a problem as the interval selection criteria $\mc{L}(c) = \frac{\mc{L}(a) + \mc{L}(b)}{2} - K \cdot (b - a)$ will be mainly determined by the interval width $b - a$. In other words, the Shubert-Piyavskii algorithm will focus on global exploration in applications on real-world problems.
 
\item \textbf{Computational Complexity}: The Shubert-Piyavskii algorithm introduced here works well for the objective function with one single variable. When it comes to the scenario with $d_{\theta}$ variables, the exploration of the Shubert-Piyavskii algorithm will involve the evaluations at $2^{d_{\theta}}$ endpoints, which will become very inefficient. To resolve such a problem, several variant version of the Shubert-Piyavskii algorithm have been proposed for the multivariate objective functions, including \cite{galperin, pinter_globally, Mladineo1986}. However, these extension algorithms may still be very computationally complex for some other reasons, which also hinders its application in many real-world optimization problems. 

\end{itemize}


\subsection{{\direct} Algorithm}\label{subsec:direct}

The Shubert-Piyavskii algorithm requires the prior knowledge of the Lipschitz constant $K$ for the objective function in advance, which may be impractical in real-world applications. In this part, we will introduce the {\direct} (DIviding RECTangles) algorithm \cite{direct}, which can also effectively resolve the two disadvantages mentioned above for the Shubert-Piyavskii algorithm. We will first introduce the {\direct} algorithm for the one-dimensional variable first, and then talk about its extensions to the multi-dimensional variable objective function.

\subsubsection{One-dimensional {\direct} Algorithm}

According to the analysis provided before, one of the great challenges that hinder the application of the Shubert-Piyavskii algorithm in multivariate objective functions is due to its huge evaluation costs, which grows exponentially with the variable dimensions. Instead of evaluating the objective function at the endpoints of an interval, the {\direct} algorithm evaluates the function at the center point instead. In other words, when it deals with the multivariate objective function (e.g., with $n$ variable), the function evaluation will need to be done just once, instead of the $2^n$ as required by the Shubert-Piyavskii algorithm.

\begin{figure}[t]
    \centering
    \includegraphics[width=0.6\textwidth]{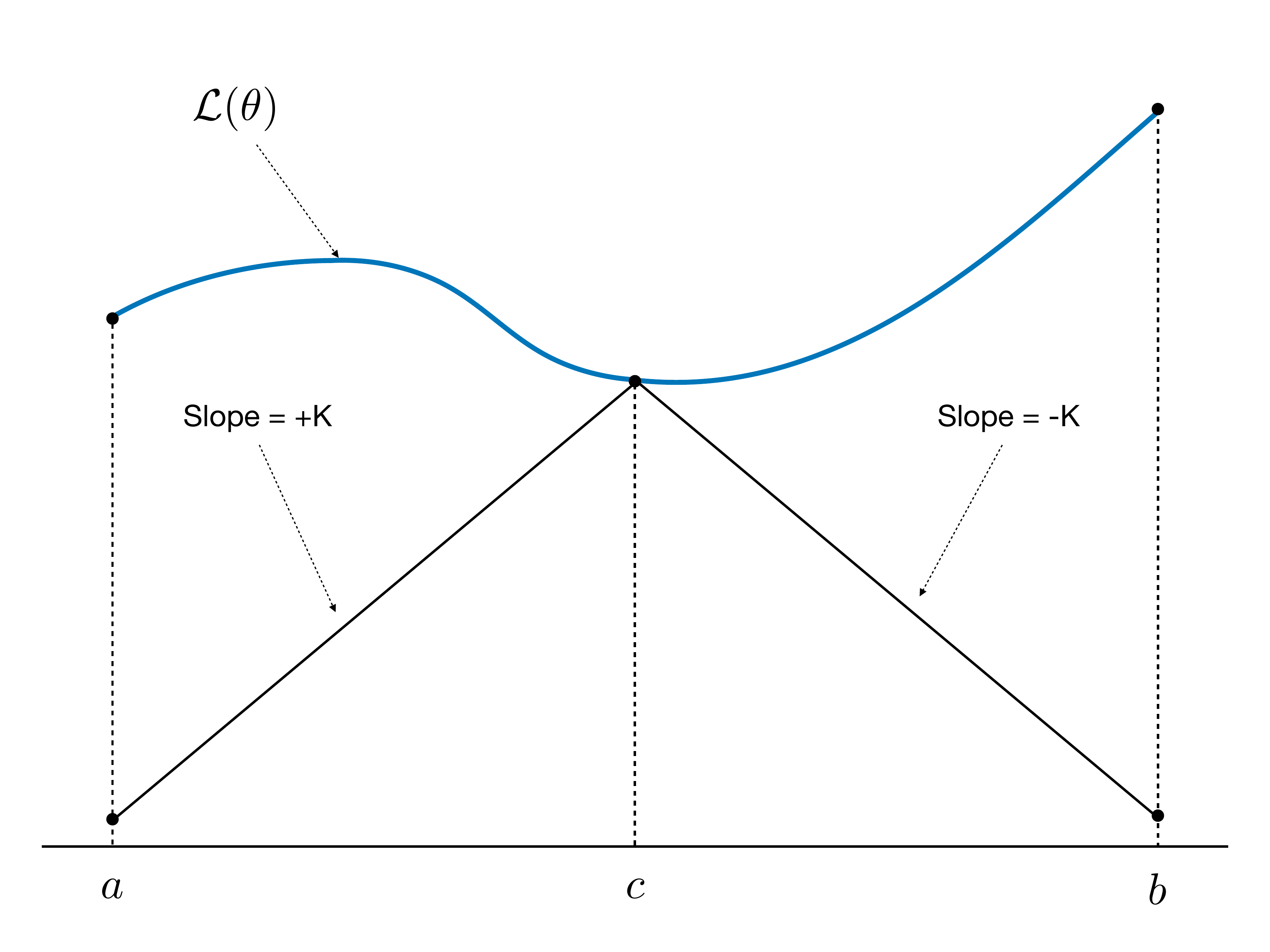}
    \caption{An Example to Illustrate the {\direct} Algorithm.}
    \label{fig:chap_derivative_free_direct}
\end{figure}

Based on the Lipschitz function (i.e., the inequality in Formula~\ref{equ:lipschitz_function}), by replacing $\theta_1$ with $\theta$ and $\theta_2$ with $c$, we can easily get the following two inequalities:
\begin{align*}
&\mc{L}(\theta) \ge \mc{L}(c) + K (\theta -c), \forall \theta \le c;\\
&\mc{L}(\theta) \ge \mc{L}(c) - K(\theta -c), \forall \theta \ge c,
\end{align*}
where $c =\frac{a+b}{2}$ denotes the center point of interval $[a, b]$. 

Here, we assume the Lipschitz constant $K$ is still known, and we will introduce the approach for unknown Lipschitz constant scenarios later. As illustrated in Figure~\ref{fig:chap_derivative_free_direct}, these two inequalities actually correspond to the two lines of slops $-K$ and $+K$ forming a ``$\Lambda$'' structure below the function curve. Values of these two lines at points $a$ and $b$ will define the lowest value of them within the interval $[a, b]$, i.e., 
\begin{equation}\label{equ:lipo_lower_bound}
\mbox{lower bound} = \mc{L}(c) - K \frac{b - a}{2}.
\end{equation}

The interval division for the {\direct} algorithm is slightly different from the Shubert-Piyavskii algorithm, which partitions the interval into $3$ sub-intervals instead. 

\begin{example}

For instance, given the interval before division in Figure~\ref{fig:chap_derivative_free_direct_division}, the {\direct} algorithm partitions it into $3$ segments as illustrated in the plot after division. Among all the potential intervals, the {\direct} algorithm will pick the interval with a smaller lower-bound (as defined in Equ~\ref{equ:lipo_lower_bound}) to explore. In many cases the strict Lipschitz constant $K$ is hard to compute in advance, and the {\direct} algorithm allows us to use a rate-of-change constant $\tilde{K}$ to replace it instead. 

\end{example}

\begin{figure}[t]
    \centering
    \includegraphics[width=0.6\textwidth]{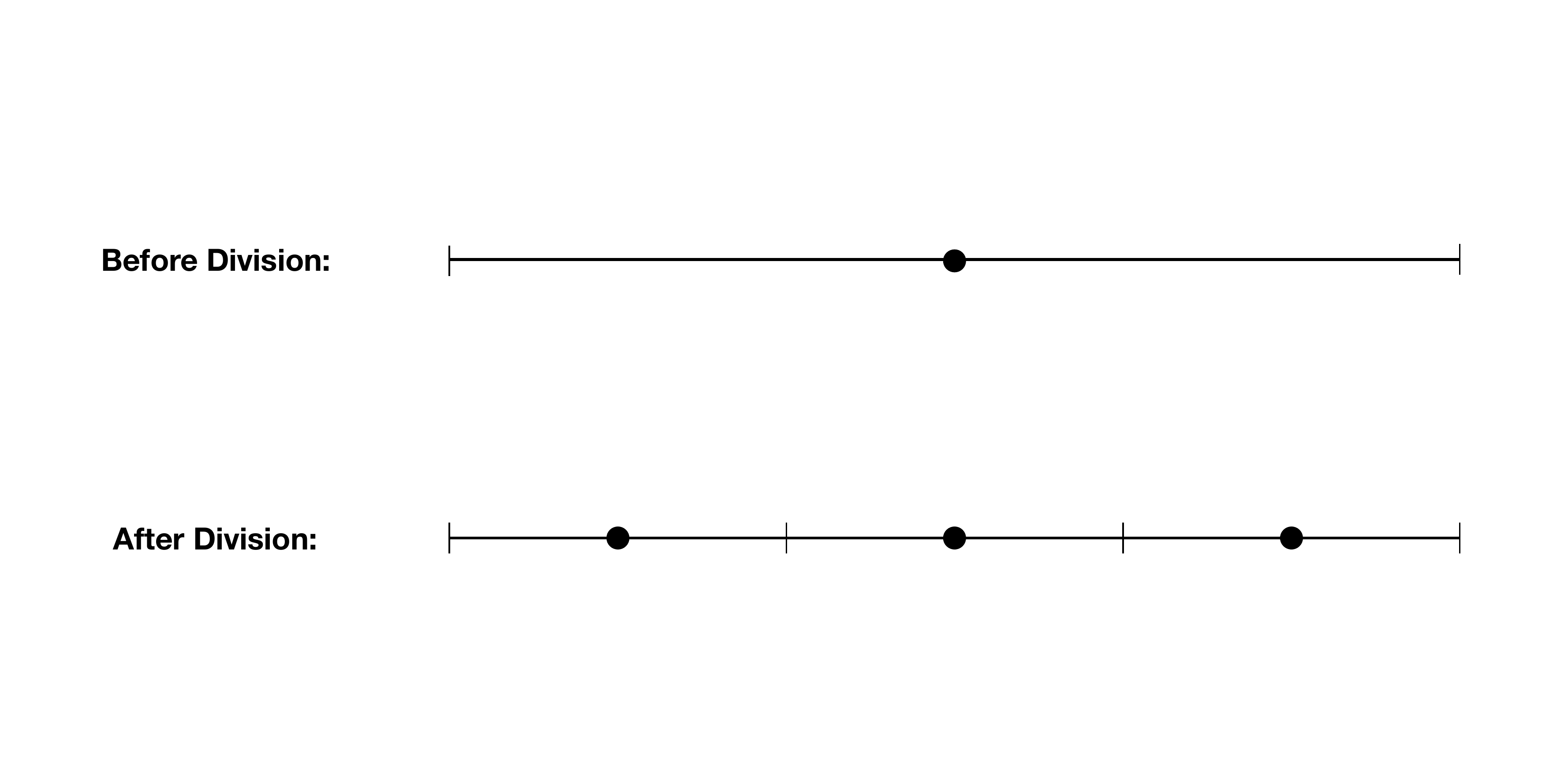}
    \caption{A Example for Division in the {\direct} Algorithm.}
    \label{fig:chap_derivative_free_direct_division}
\end{figure}

Suppose we have partition the original interval $[a, b]$ into a group of sub-intervals $\{[a_i, b_i]\}_{i =1, 2, \cdots, m}$ with the corresponding midpoints $\{c_i\}_{i=1, 2, \cdots, m}$. Let $\mc{L}_{min}$ denote the current best function value and $\epsilon$ denote a positive constant. Interval $[a_j, b_j], j \in \{1, 2, \cdots, m\}$ will be selected selected to explore next iff there exists some rate-of-change constant $\tilde{K}$ such that
\begin{align}\label{equ:direct_division_interval_selection_one_dimensional}
&\mc{L}(c_j) - \tilde{K}\frac{b_j - a_j}{2} \le \mc{L}(c_i) - \tilde{K}\frac{b_i - a_i}{2}, \forall i = 1, 2, \cdots, m;\\
&\mc{L}(c_j) - \tilde{K}\frac{b_j - a_j}{2} \le \mc{L}_{min} - \epsilon | \mc{L}_{min} |.
\end{align}
For the above inequalities, the first one indicates that interval $[a_j, b_j]$ to be on the lower right of the convex hull of the dots, and the second condition requires the selected interval to be better than the current best solution by a nontrivial amount. The pseudo-code of the one-dimensional {\direct} algorithm is provided in Algorithm~\ref{alg:one_dimensional_direct_algorithm}.

\begin{algorithm}[t]
\small
\caption{One-Dimensional {\direct} Algorithm}
\label{alg:one_dimensional_direct_algorithm}
\begin{algorithmic}[1]
	\REQUIRE Variable range [a, b]; Rate-of-change parameter $\tilde{K}$; Constant $\epsilon$.
\ENSURE  Model Parameter ${\theta}$
\STATE	{Evaluate $\mc{L}(c)$ at the midpoint of intervan $[a, b]$.}
\STATE	{Initialize set $\mc{S} = \{[\mc{L}(c): (a, b; c)]\}$.}
\STATE	{Initialize $\mc{L}_{min} = \mc{L}(c)$.}
\WHILE	{Stop criteria is not met.}
\STATE	{Select the optimal $[a_i, b_i]$ according to Inequalities~(\ref{equ:direct_division_interval_selection_one_dimensional}).}
\STATE	{Compute $\mc{S}$ size $m = |\mc{S}|$.}
\STATE	{$\mc{S} = \mc{S} \setminus \{[\mc{L}(c_i): (a_i, b_i; c_i)]\}$.}
\STATE	{Compute $\delta = (b_i - a_i)/3$.}
\STATE	{Update the interval $[a_{i}, b_{i}] = [a_i+\delta, a_i+2\delta]$.}
\STATE	{Create two new intervals $[a_{m+1}, b_{m+1}] = [a_i, a_i + \delta]$ and $[a_{m+2}, b_{m+2}] = [a_i + 2\delta, b_i]$.}
\STATE	{Evaluate $\mc{L}(c_{m+1})$ and $\mc{L}(c_{m+2})$ at the midpoints $c_{m+1}$ and $c_{m+2}$.}
\STATE	{Update $\mc{L}_{min} = \min\left(\mc{L}_{min}, \mc{L}(c_{m+1}), \mc{L}(c_{m+2}) \right)$.}
\STATE	{Update set with the updated/new intervals $\mc{S} = \mc{S} \cup \{[\mc{L}(c_i): (a_i, b_i; c_i)], [\mc{L}(c_{m+1}): (a_{m+1}, b_{m+1}; c_{m+1})], [\mc{L}(c_{m+2}): (a_{m+2}, b_{m+2}; c_{m+2})]\}$.}
\ENDWHILE
\end{algorithmic}
\end{algorithm}

\subsubsection{Multi-dimensional {\direct} Algorithm}

In this part, we will introduce the multi-dimensional {\direct} algorithm for computing the solution to the multivariate objective function. To simplify the problem settings, we assume every variable is within a pre-specified bound $[0, 1]$, which can be achieved by variable rescaling easily. In other words, the search space of the problem with be a $n$-dimensional unit hyper-cube instead, which will be partitioned into hyper-rectangles by the {\direct} algorithm in the learning process with a sample point in each of them. The main challenge to be studied here will be how to partition the hyper-cube for the multi-dimensional {\direct} algorithm.

The division process in the multi-dimensional {\direct} Algorithm is different from the division process in the uni-dimensional {\direct} Algorithm. Formally, let $\mb{c}$ denote the center point of the current hyper-cube to be explored, whose dimension length can be denoted as $3\delta$ (i.e., one-third of the dimension length will be $\delta$). Therefore, the potential points to be evaluated here can be denoted as $\{\mb{c} \pm \delta \cdot \mb{e}_i\}_{i = 1}^{n}$, where vector $\mb{e}_i$ contains $1$ at the $i_{th}$ entry and $0$s at the remaining entries.

\begin{figure}[t]
    \centering
    \includegraphics[width=0.6\textwidth]{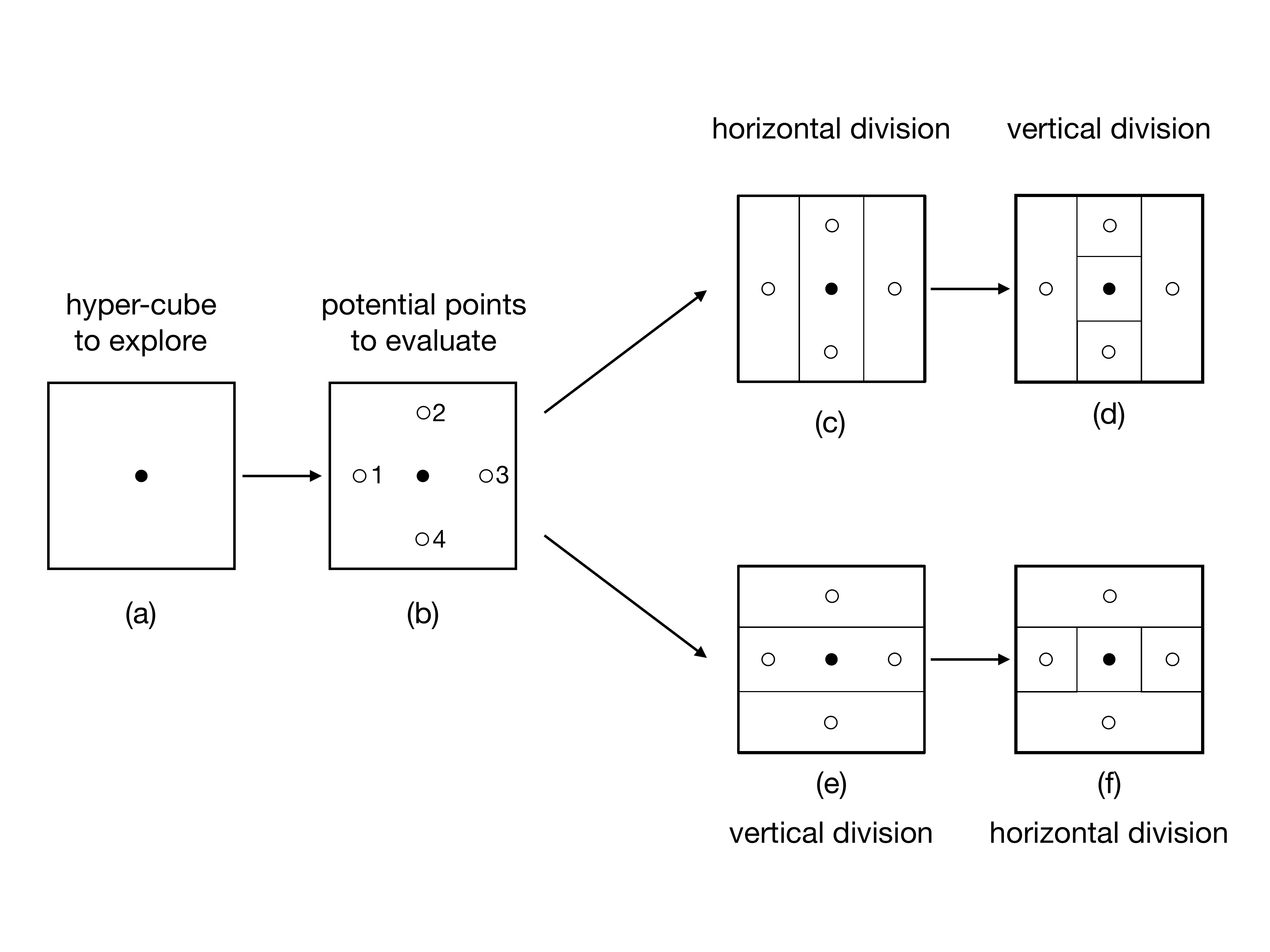}
    \caption{A Example of {\direct} in 2D Division.}
    \label{fig:chap_derivative_free_direct_division}
\end{figure}

\begin{example}

For instance, we provide an example in Figure~\ref{fig:chap_derivative_free_direct_division}, where plot (a) is the hyper-cube to be explored. Based on the dimension length as well as the central solid point, we can represent the set of points to be evaluated as set $\{1, 2, 3, 4\}$, i.e., the $4$ hollow dots in plot (b). Dots $1$ and $3$ are the ones sampled from the horizontal dimension, while dots $2$ and $4$ are sampled from the vertical dimension instead.

\end{example}

Figure~\ref{fig:chap_derivative_free_direct_division} also illustrate two different ways to partition the hyper-cube into smaller regions:
\begin{enumerate}
\item Plots (c) and (d) illustrate the partition of the hyper-cube in the horizontal dimension first, and then partition it along the vertical dimension.
\item Plots (e) and (f) illustrate the partition of the hyper-cube along the vertical dimension first, and then along the horizontal dimension.
\end{enumerate}

To determine which division is more promising, the strategy adopted in the {\direct} algorithm is to make the biggest rectangles contain the best function values. Among all the potential dimensions $i = \{1, 2, \cdots, n\}$, the {\direct} algorithm will compute a weight $w_i$ for each dimension as follows:
\begin{equation}\label{equ:multi_dimension_direct_weight}
w_i = \min \{\mc{L}(\mb{c} + \delta \mb{e}_i), \mc{L}(\mb{c} - \delta \mb{e}_i)\},
\end{equation}
which denotes the best function value along dimension $i$.

The optimal dimension $i^*$ will be selected to divide based on the weight values $\{w_i\}_{i = 1, 2, \cdots, n}$, i.e.,
\begin{equation}
i^* = \arg \min_{i \in \{1, 2, \cdots, n\}} w_i.
\end{equation}

Meanwhile, the selection of the optimal hyper-rectangles for the multivariate {\direct} algorithm is very similar to the one-dimensional version. Formally, let's assume we have partition the original hyper-cube into $m$ hyper-rectangles and we can denote all the current hyper-rectangles available with their centers, i.e., $\{\mb{c}_i\}_{i = 1, 2, \cdots, m}$. A hyper-rectangle centered at $\mb{c}_j$ is said to be optimal if there exists some $\tilde{K}$ such that
\begin{align}\label{equ:direct_division_interval_selection}
&\mc{L}(c_j) - \tilde{K}{d}_j \le \mc{L}(c_i) - \tilde{K}{d}_i, \forall i = 1, 2, \cdots, m;\\
&\mc{L}(c_j) - \tilde{K}{d}_j \le \mc{L}_{min} - \epsilon | \mc{L}_{min} |,
\end{align}
where $d_j$ denotes the distance from the center $\mb{c}_j$ to the vertices of the $j_{th}$ hyper-rectangle.

Based on the above descriptions, we can provide the pseudo-code of multivariate {\direct} in Algorithm~\ref{alg:multi_dimensional_direct_algorithm}. As indicated in \cite{direct}, the {\direct} algorithm is guaranteed to converge to the global optimal if the objective function is continuous, or at least continuous in the neighborhood of a global optimum.

\begin{algorithm}[t]
\small
\caption{Multi-Dimensional {\direct} Algorithm}
\label{alg:multi_dimensional_direct_algorithm}
\begin{algorithmic}[1]
	\REQUIRE Variable range hyper-rectangle; Rate-of-change parameter $\tilde{K}$; Constant $\epsilon$.
\ENSURE  Model Parameter $\boldsymbol{\theta}$
\STATE	{Normalize the search space to be a unit hyper-cube.}
\STATE	{Evaluate $\mc{L}(\mb{c})$ at the center point of the hyper-cube.}
\STATE	{Initialize set $\mc{S} = \{[\mc{L}(\mb{c}): \mb{c}]\}$.}
\STATE	{Initialize $\mc{L}_{min} = \mc{L}(\mb{c})$.}
\STATE	{Initialize hyper-rectangle number $m = 1$.}
\WHILE	{stop criteria is not met.}
\STATE	{Identify the set of optimal hyper-rectangles $\mc{S}' \subset \mc{S}$.}
\FOR	{hyper-rectangle centered at $\mb{c}_j$ in $\mc{S}'$}
\STATE	{Identify the set $\mc{I}$ of dimensions of the maximum side length.}
\STATE	{Evaluate the function at points $\{\mb{c} \pm \mb{e}_i\}_{i \in \mc{I}}$.}
\STATE	{Sort the dimensions in $\mc{I}$ according to the weights $w_i = \min \{\mc{L}(\mb{c} + \delta \mb{e}_i), \mc{L}(\mb{c} - \delta \mb{e}_i)\}$ for dimension $i$ computed according to Equation~\ref{equ:multi_dimension_direct_weight}.}
\STATE	{Divide the hyper-rectangle containing $\mb{c}$ into thirds along the dimensions in $\mc{I}$, starting with the dimension with the lowest $w_i$ and the so forth.}
\ENDFOR
\STATE	{Update $\mc{L}_{min}$ and $m$ according to the points in the newly sampled hyper-rectangles.}
\ENDWHILE
\end{algorithmic}
\end{algorithm}


\subsection{LIPO Algorithm}

The global optimization algorithms we have introduced in this section, including both the Shubert-Piyavskii algorithm and the {\direct} algorithm, is shown to converge subject to the local smoothness assumption. However, by this context so far, such convergence properties of these algorithms have not been considered in the scenarios where only the global smoothness assumption on the function can be specified. In this part, we will introduce another Lipschitz function based global optimization algorithm, namely {\lipo} \cite{lipo}. {\lipo} effectively exploits the global smoothness of the unknown function, and is shown to converge faster on globally smooth problems than the other existing Lipschitz function based global optimization algorithms.

\subsubsection{LIPO with Known Lipschitz Constant}

Similar to the Shubert-Piyavskii algorithm and the {\direct} algorithm, {\lipo} is also a sequential algorithm, whose optimization process involves a sequence of sampled variable points and the function evaluations at these points. Formally, given the variable search space $\Theta$, we can represent these sampled points as a sequence $\left(\boldsymbol{\theta}_1, \boldsymbol{\theta}_2, \cdots, \boldsymbol{\theta}_n \right)$, where $\boldsymbol{\theta}_i$ denotes the point uniformly sampled from $\Theta$ at iteration $i \in \{1, 2, \cdots, n\}$. The {\lipo} algorithm will decide whether or not to evaluate the function at the point subject to certain criteria. 

Before we talk about the evaluation criteria, we need to introduce several important concepts as follows, including \textit{consistent function} and \textit{potential maximizer}. Formally, give the Lipschitz constant $K \ge 0$, we can represent the set of functions subject to the corresponding Lipschitz function inequality (i.e., Formula~\ref{equ:lipschitz_function}) as $Lip(K)$, namely the $K$-Lipschitz functions. Based on the sequence of sampled points together with the function evaluations, a subset of the functions in $Lip(K)$ can be identified as the \textit{consistent functions}. 

\begin{definition}
(\textit{Consistent Functions}): The active subset of the $K$-Lipschitz functions consistent with the unknown function $\mc{L}(\cdot)$ over a sequence of $n$ sampled points and function evaluations $[(\boldsymbol{\theta}_1, \mc{L}(\boldsymbol{\theta}_1) ), (\boldsymbol{\theta}_2, \mc{L}(\boldsymbol{\theta}_2) ), \cdots, (\boldsymbol{\theta}_n, \mc{L}(\boldsymbol{\theta}_n) )]$ ($n \ge 1$) is defined as follows
\begin{equation}
\mc{F}_{K, n} = \left\{ g \in Lip(K): \forall i \in \{1, 2, \cdots, n\}, g(\boldsymbol{\theta}_i)=\mc{L}(\boldsymbol{\theta}_i) \right\}.
\end{equation}
\end{definition}

Based on the above definition, the functions in set $\mc{F}_{K, n}$ can achieve consistent values with the evaluation of $\mc{L}(\cdot)$ at these sampled points in the search space. Considering that function $\mc{L}(\cdot)$ is unknown, its optimal variable can be potentially identified with the function candidates in set $\mc{F}_{K, n}$.

\begin{definition}
(\textit{Potential Maximizers}): Given the sampled points and function evaluation sequence $[(\boldsymbol{\theta}_1, \mc{L}(\boldsymbol{\theta}_1) ), (\boldsymbol{\theta}_2, \mc{L}(\boldsymbol{\theta}_2) ), \cdots, (\boldsymbol{\theta}_n, \mc{L}(\boldsymbol{\theta}_n) )]$ together with the set of \textit{consistent functions} $\mc{F}_{K, n}$, we can represent the set of potential maximizers as
\begin{equation}
{\Theta}_{K, n} = \left\{\boldsymbol{\theta} \in \Theta: \exists g \in \mc{F}_{K, n} \mbox{ such that } \boldsymbol{\theta} \in \arg \max_{\boldsymbol{\theta}' \in \Theta} g(\boldsymbol{\theta}') \right\}.
\end{equation}
\end{definition}

The set ${\Theta}_{K, n}$ defines the space that function $\mc{L}(\cdot)$ can achieve the maximum values. Given any variable $\boldsymbol{\theta}$, the notation ``$\boldsymbol{\theta} \in {\Theta}_{K, n}$'' has an equivalent representation according to the following Lemma~\ref{lemma:lipo_criteria}.
\begin{lemma}\label{lemma:lipo_criteria}
Let ${\Theta}_{K, n}$ denote the set of potential maximizer defined above, we can achieve the following equivalent representation for any variable $\boldsymbol{\theta} \in {\Theta}_{K, n}$:
\begin{equation}
\boldsymbol{\theta} \in {\Theta}_{K, n} \Leftrightarrow \min_{i \in \{1, 2, \cdots, n\}} \left( \mc{L}(\boldsymbol{\theta}_i) + K \cdot \left\|\boldsymbol{\theta} - \boldsymbol{\theta}_i \right\|_2 \right) \ge \max_{i' \in \{1, 2, \cdots, n\}} \mc{L}(\boldsymbol{\theta}_{i'}).
\end{equation}
\end{lemma} 

\begin{algorithm}[t]
\small
\caption{{\lipo} Algorithm}
\label{alg:lipo_known}
\begin{algorithmic}[1]
	\REQUIRE Variable search space $\Theta$; Lipschitz constant ${K}$; Constant $n$.
\ENSURE  Model Parameter $\boldsymbol{\theta}$
\STATE	{Initialization: Sample $\boldsymbol{\theta}_1 \sim \mc{U}(\Theta)$ from $\Theta$ uniformly.}
\STATE	{Evaluate function value $\mc{L}(\boldsymbol{\theta}_1)$.}
\STATE	{Add $(\boldsymbol{\theta}_1, \mc{L}(\boldsymbol{\theta}_1))$ to the sampling sequence.}
\STATE	{Initialize index tag $t = 2$}
\WHILE	{$t \le n$}
\STATE	{Sample $\boldsymbol{\theta} \sim \mc{U}(\Theta)$ from $\Theta$ uniformly.}
\IF		{$\min_{i \in \{1, 2, \cdots, t\}} \left( \mc{L}(\boldsymbol{\theta}_i) + K \cdot \left\|\boldsymbol{\theta} - \boldsymbol{\theta}_i \right\|_2 \right) \ge \max_{i' \in \{1, 2, \cdots, t\}} \mc{L}(\boldsymbol{\theta}_{i'})$}
\STATE	{Assign $\boldsymbol{\theta}_t = \boldsymbol{\theta}$.}
\STATE	{Evaluate function $\mc{L}(\boldsymbol{\theta}_t)$.}
\STATE	{Add $(\boldsymbol{\theta}_t, \mc{L}(\boldsymbol{\theta}_t))$ to the sampling sequence.}
\STATE	{$t=t+1$}
\ENDIF
\ENDWHILE
\STATE	{Return $\boldsymbol{\theta}^* = \arg \max_{i \in \{1, 2, \cdots, n\}} \mc{L}(\boldsymbol{\theta}_i)$}
\end{algorithmic}
\end{algorithm}

\begin{proof}
We will prove the above Lemma from two directions:
\begin{itemize}
\item \textit{Direction 1}: $\boldsymbol{\theta} \in {\Theta}_{K, n} \Rightarrow \min_{i \in \{1, 2, \cdots, n\}} \left( \mc{L}(\boldsymbol{\theta}_i) + K \cdot \left\|\boldsymbol{\theta} - \boldsymbol{\theta}_i \right\|_2 \right) \ge \max_{i' \in \{1, 2, \cdots, n\}} \mc{L}(\boldsymbol{\theta}_{i'})$.

Given that variable $\boldsymbol{\theta} \in {\Theta}_{K, n}$, we can have $\boldsymbol{\theta} \in \arg \max_{\boldsymbol{\theta}' \in \Theta} g(\boldsymbol{\theta}')$ for some function $g \in \mc{F}_{K, n}$. Considering that $g$ is a $K$-Lipschitz function, we can have 
\begin{equation}\label{equ:lip_lemma_prove_direction_1}
\left| g(\boldsymbol{\theta}_1) - g(\boldsymbol{\theta}_2) \right| \le K \cdot \left\| \boldsymbol{\theta}_1 - \boldsymbol{\theta}_2 \right\|_2, \forall \boldsymbol{\theta}_1, \boldsymbol{\theta}_2 \in \Theta.
\end{equation}
Here, let's assign $\boldsymbol{\theta}_1 = \arg \max_{\boldsymbol{\theta}' \in \Theta} g(\boldsymbol{\theta}')$ and $\boldsymbol{\theta}_2 = \arg \min_{i \in \{1, 2, \cdots, n\}} \left( g(\boldsymbol{\theta}_i) + K \cdot \left\| \boldsymbol{\theta}_1 - \boldsymbol{\theta}_i \right\|_2 \right)$, we have
\begin{equation}
g(\boldsymbol{\theta}_1) \ge g(\boldsymbol{\theta}_2) 
\end{equation}
and according to Inequality~\ref{equ:lip_lemma_prove_direction_1}, we can get 
\begin{equation}
\max_{i' \in \{1, 2, \cdots, n\}} g(\boldsymbol{\theta}_{i'}) \le g(\boldsymbol{\theta}_1) \le g(\boldsymbol{\theta}_2) + K \cdot \left\| \boldsymbol{\theta}_1 - \boldsymbol{\theta}_2 \right\|_2.
\end{equation}
Considering that function $g \in \mc{F}_{K, n}$, function evaluation $g(\boldsymbol{\theta}_i) = \mc{L}(\boldsymbol{\theta}_i), \forall i \in \{1, 2, \cdots, n\}$. In other words, we have
\begin{equation}
\max_{i' \in \{1, 2, \cdots, n\}} \mc{L}(\boldsymbol{\theta}_{i'}) \le \min_{i \in \{1, 2, \cdots, n\}} \mc{L}(\boldsymbol{\theta}_{i}) + K \cdot \left\| \boldsymbol{\theta}_1 - \boldsymbol{\theta}_{i} \right\|_2,
\end{equation}
where $\boldsymbol{\theta}_1 \in {\Theta}_{K, n}$.

\item \textit{Direction 2}: $\min_{i \in \{1, 2, \cdots, n\}} \left( \mc{L}(\boldsymbol{\theta}_i) + K \cdot \left\|\boldsymbol{\theta} - \boldsymbol{\theta}_i \right\|_2 \right) \ge \max_{i' \in \{1, 2, \cdots, n\}} \mc{L}(\boldsymbol{\theta}_{i'}) \Rightarrow \boldsymbol{\theta} \in {\Theta}_{K, n}$.
According to the previous analysis, we know that for any $\boldsymbol{\theta} \in \Theta$, we have
\begin{equation}
\mc{L}(\boldsymbol{\theta}) \le  \min_{i \in \{1, 2, \cdots, n\}} \left( \mc{L}(\boldsymbol{\theta}_i) + K \cdot \left\|\boldsymbol{\theta} - \boldsymbol{\theta}_i \right\|_2 \right).
\end{equation}

In other words, the terms on the right-hand-side actually defines the upper bound of potential values that function $\mc{L}(\cdot)$ can achieve at point $\boldsymbol{\theta}$, which can be formally defined as follows:
\begin{equation}
UB(\boldsymbol{\theta}) = \min_{i \in \{1, 2, \cdots, n\}} \left( \mc{L}(\boldsymbol{\theta}_i) + K \cdot \left\|\boldsymbol{\theta} - \boldsymbol{\theta}_i \right\|_2 \right).
\end{equation}

Meanwhile, the space denoted by the following inequality actually defines the set of potential variable $\boldsymbol{\theta}$ which can achieve the maximum value for the function $\mc{L}(\cdot)$:
\begin{equation}
\Theta_{\textsc{ub}} = \left\{ \boldsymbol{\theta}: \boldsymbol{\theta} \in \Theta, UB(\boldsymbol{\theta}) \ge \max_{i' \in \{1, 2, \cdots, n\}} \mc{L}(\boldsymbol{\theta}_{i'}) \right\}.
\end{equation}

Therefore, $\forall \boldsymbol{\theta} \notin \Theta_{K,n}$, we can have $\boldsymbol{\theta} \notin \Theta_{\textsc{ub}}$ (i.e., $\forall \boldsymbol{\theta} \in \Theta_{\textsc{ub}} \Rightarrow \boldsymbol{\theta} \in \Theta_{K,n}$), which proves the Direction 2.
\end{itemize}
\end{proof}

The inequality $\min_{i \in \{1, 2, \cdots, n\}} \left( \mc{L}(\boldsymbol{\theta}_i) + K \cdot \left\|\boldsymbol{\theta} - \boldsymbol{\theta}_i \right\|_2 \right) \ge \max_{i' \in \{1, 2, \cdots, n\}} \mc{L}(\boldsymbol{\theta}_{i'})$ mentioned in the above Lemma~\ref{lemma:lipo_criteria} can serve as the criteria for function evaluation in the sampling process of {\lipo}, and the pseudo-code of {\lipo} in Algorithm~\ref{alg:lipo_known}.

\subsubsection{AdaLIPO with Unknown Lipschitz Constant}

In many cases, the Lipschitz constant is hard to know in advance. In this part we will introduce the {\adalipo} (Adaptive {\lipo}) algorithm for optimizing function $\mc{L} \in \cup_{k\ge0} Lip(k)$, whose specific Lipschitz constant is unknown.

The pseudo-code of {\adalipo} is provided in Algorithm~\ref{alg:adalipo_unknown}. Compared with {\lipo}, {\adalipo} don't need unknown Lipschitz constant $K$, but will take two more parameters: (1) Bernoulli distribution parameter $p$, and (2) Lipschitz constant meshgrid $K_{i \in \mathbbm{Z}}$. Here, the notation $K_{i \in \mathbbm{Z}}$ defines a meshgrid of $\mathbbm{R}^+$ such that $\forall x > 0, \exists i \in \mathbbm{Z}$ with $K_i \le x \le K_{i+1}$. The {\adalipo} algorithm starts with an initialized Lipschitz constant $\hat{K} = 0$, and proceeds with the exploration by sampling the variables from either $\Theta$ or $\Theta_{\hat{K}_t, t}$, which is controlled via a Bernoulli distribution variable $B_{t} \sim \mc{B}(p)$. According to the algorithm pseudo-code, {\adalipo} extends {\lipo} by adding a step to infer the potential Lipschitz constant $\hat{K}_t$ (i.e., Line 16 in Algorithm~\ref{alg:adalipo_unknown}) in each step iteratively. 

Besides the inference technique introduced in Algorithm~\ref{alg:adalipo_unknown}, some other Lipschitz constant inference approaches proposed in \cite{lip_constant_inference1, lip_constant_inference2} can be used as well to implement the algorithm. To be consistent with \cite{lipo}, we will not introduce them here and the readers may refer to the cited papers for more information.

\begin{algorithm}[t]
\small
\caption{{\adalipo} Algorithm}
\label{alg:adalipo_unknown}
\begin{algorithmic}[1]
	\REQUIRE Variable search space $\Theta$; Constant $n$; Bernoulli distribution parameter $p$, Lipschitz constant meshgrid $K_{i \in \mathbbm{Z}}$.
\ENSURE  Model Parameter $\boldsymbol{\theta}$
\STATE	{Initialization: Sample $\boldsymbol{\theta}_1 \sim \mc{U}(\Theta)$ from $\Theta$ uniformly.}
\STATE	{Evaluate function value $\mc{L}(\boldsymbol{\theta}_1)$.}
\STATE	{Add $(\boldsymbol{\theta}_1, \mc{L}(\boldsymbol{\theta}_1))$ to the sampling sequence.}
\STATE	{Initialize inferred Lipschitz constant $\hat{K}_{2} = 0$.}
\STATE	{Initialize index tag $t = 2$}
\WHILE	{$t \le n$}
\STATE	{Sample variable $B_{t} \sim \mc{B}(p)$ from the Bernoulli distribution.}
\IF		{$B_t = 1$}
\STATE	{Sample $\boldsymbol{\theta}_t \sim \mc{U}(\Theta)$ from $\Theta$.}
\ELSE	
\STATE	{Sample $\boldsymbol{\theta}_t \sim \mc{U}(\Theta_{\hat{K}_t, t})$ from $\Theta_{\hat{K}_t, t}$ (which denotes the set of potential optimizers computed with $\hat{K}_t$ and $t$).}
\ENDIF	
\STATE	{Evaluate function $\mc{L}(\boldsymbol{\theta}_t)$.}
\STATE	{Add $(\boldsymbol{\theta}_t, \mc{L}(\boldsymbol{\theta}_t))$ to the sampling sequence.}
\STATE	{$t=t+1$}
\STATE	{Update $\hat{K}_i = \inf \left\{K_{i \in \mathbbm{Z}}: \max_{i \neq j} \frac{|\mc{L}(\boldsymbol{\theta}_i) - \mc{L}(\boldsymbol{\theta}_j) |}{\left\| \boldsymbol{\theta}_i - \boldsymbol{\theta}_j \right\|_2} \le K_i \right\}$.}
\ENDWHILE
\STATE	{Return $\boldsymbol{\theta}^* = \arg \max_{i \in \{1, 2, \cdots, n\}} \mc{L}(\boldsymbol{\theta}_i)$}
\end{algorithmic}
\end{algorithm}

\subsection{{\mcs} Algorithm}

The {\mcs} algorithm \cite{mcs} to be introduced here is a heuristic-based optimization algorithm, and it has a close relationship with the {\direct} algorithm introduced before. In {\direct}, the search space will be normalized to a hyper-cube $[0, 1]^{d_{\theta}}$, which will be partitioned into smaller hyper-rectangles in the searching process. Each hyper-rectangle is characterized by its center-point, and the dimension length of the hyper-rectangles will be always $3^{-k}, k \in \mathbbm{N}$. The {\direct} algorithm may suffer from several disadvantages: (1) {\direct} cannot handle the search space with infinite bounds; and (2) {\direct} cannot reach the boundary and may lead to slow learning process if the minimum lies at the boundaries. The {\mcs} algorithm to be introduced in this part can resolve such these problems.

\subsubsection{Overview of {\mcs}}

Here, we will provide the overview of {\mcs}, whose detailed information will be provided as follows. Similar to the Shubert-Piyavskii algorithm and {\direct}, {\mcs} also combines the global search with local search, where ``global search'' denotes the division of hyper-rectangles with large unexplored territory and ``local search'' denotes the division of hyper-rectangles with the optimal function values (corresponding to the two terms indicated in Equations~\ref{equ:direct_division_interval_selection} and \ref{equ:lipo_lower_bound}). {\mcs} adopts a multi-level search strategy to balance between global search and local search, where each hyper-rectangle will be assigned with a level index $s \in \{0, 1, \cdots, s_{max}\}$. Here, level $s=0$ denotes the hyper-rectangle has been divided and will be ignored for further division, and level $s=s_{max}$ denotes the hyper-rectangle is too small for further division. Meanwhile, for the hyper-rectangles with a level $0 < s < s_{max}$, after division, their level will be set to be $s=0$ but their descendants will get level $s+1$ or $\min(s+2, s_{max})$. Viewed in such a perspective, the hyper-rectangle level $s$ will denote its size effectively, and the ones with a small level will have a larger size (except level $0$).

After the initialization step, the {\mcs} algorithm will \textit{sweep} through all the available hyper-rectangles with levels $0 < s < s_{max}$ from the lowest levels to the higher ones (which corresponds to the global part in {\mcs}). For the hyper-rectangles belonging to the same level, {\mcs} will select the ones with the lowest function evaluation values, which corresponds to the local part in {\mcs}. For the finally selected hyper-rectangles, {\mcs} will divide them along one single coordinate (i.e., one single variable) in one step, and the coordinate as well as the division point are determined by the information from the sampled points. Different from {\direct}, the division points lies at the boundaries of the hyper-rectangle in {\mcs}, which allows {\mcs} to resolve the second disadvantage of {\direct} as mentioned before. With multiple times of division, the hyper-rectangles will become arbitrarily small and the hyper-rectangles will shrink sufficiently fast.

{\mcs} without local search will put the sampled points together with their function evaluations of hyper-rectangles of level $s_{max}$ into a basket (containing the ``useful'' points). Meanwhile, the {\mcs} with local search will start a local search from these points before putting them into the basket so as to accelerate the convergence of the algorithm. {\mcs} with local search will first check whether the point in the current hyper-rectangle is in the basin of attraction of a local minimizer or not. If it is not the case, {\mcs} will build a quadratic model by triple searches, and identify minimum of the quadratic function with a line search along a promising direction.

For the terms and techniques mention above, we will provide a much more detailed description in the following subsections in great detail.

\subsubsection{Initialization}

In {\mcs}, the sampled points lie at the boundaries of the hyper-rectangles, which may belong to multiple hyper-rectangles (which share the same boundary or vertices). These sampled points are also called the \textit{base points}. In {\mcs}, besides these base points, each hyper-rectangle is also assigned with an \textit{opposite point}, which together with the corresponding \textit{base point} can precisely identify the hyper-rectangle. Let the initial point in the hyper-rectangle as $\bs{\theta}^0$. {\mcs} starts the initialization step by selecting a set of initial base points (at least three points) along some coordinate, e.g., coordinate $i \in \{1,2, \cdots, d_{\theta}\}$ corresponding to variable $\boldsymbol{\theta}_i$, which can be denoted as $\{\boldsymbol{\theta}^1, \boldsymbol{\theta}^2, \cdots, \boldsymbol{\theta}^{l_i}\}$. These points share the same coordinates with the initial point $\bs{\theta}^0$ for all the other coordinates except the $i_{th}$ coordinate, and their $i_{th}$ coordinate takes the following values respectively
\begin{equation}
a_i \le \boldsymbol{\theta}_i^1 < \boldsymbol{\theta}_i^2 < \cdots < \boldsymbol{\theta}_i^{l_i} \le b_i,
\end{equation}
where $[a_i, b_i]$ denote the lower/upper bounds along the $i_{th}$ dimension in the search space. The objective function will be evaluated at these selected points, which can form the pairs $\left\{\left(\bs{\theta}^1, \mc{L}(\bs{\theta}^1) \right), \left(\bs{\theta}^2, \mc{L}(\bs{\theta}^2) \right), \cdots, \left(\bs{\theta}^{l_i}, \mc{L}(\bs{\theta}^{l_i}) \right) \right\}$. Among these points, the one achieving the optimal function evaluation will be denoted as $\bs{\theta}^*$.

Besides these $l_i$ points, an extra group of $l_i - 1$ division points will be selected, which can be denoted as $\{{\mb{z}}^2, {\mb{z}}^3, \cdots, {\mb{z}}^{l_1}\}$. These points also share the same coordinates with $\bs{\theta}^0$ except the $i_{th}$ dimension, which take value
\begin{equation}
{\mb{z}}^l_i = {\bs{\theta}}^{l-1}_i + q^k \left( {\bs{\theta}}^{l}_i - {\bs{\theta}}^{l-1}_i\right), l \in\{2, 3, \cdots, l_i\}.
\end{equation}
Here, in the above equation term $q = \frac{\sqrt{5}-1}{2}$ denotes the golden section ratio, and $k \in \{1, 2\}$ is chose to ensure the parts with smaller function evaluations can get larger intervals. 

Depending how many points in $\{\boldsymbol{\theta}^1, \boldsymbol{\theta}^2, \cdots, \boldsymbol{\theta}^{l_i}\}$ lie at the lower/upper boundaries, these points together with the above extra division points $\{\mb{z}^2, \mb{z}^3, \cdots, \mb{z}^{l_i}\}$ will divide the search space into $2l_i-2$ (if $a_i = \boldsymbol{\theta}_i^1$ and $\boldsymbol{\theta}_i^{l_i} = b_i$), $2l_i-1$ (if $a_i = \boldsymbol{\theta}_i^1$ or $\boldsymbol{\theta}_i^{l_i} = b_i$), and $2l_i$ (if $a_i < \boldsymbol{\theta}_i^1$ and $\boldsymbol{\theta}_i^{l_i} < b_i$) sub-hyper-rectangles. In addition, $\{\boldsymbol{\theta}^1, \boldsymbol{\theta}^2, \cdots, \boldsymbol{\theta}^{l_i}\}$ will serve as the base points of these new hyper-rectangles, which will also be used to update $\bs{\theta}^*$. If $\bs{\theta}^*$ is not shared by multiple hyper-rectangles, then the one containing $\bs{\theta}^*$ will be picked as the ``current'' hyper-rectangle for the next round of division. Otherwise, {\mcs} will select the sub-hyper-rectangle to divide for the next dimension in the initialization with a quadratic interpolation. By sampling $3$ random consecutive points selected from $\{\boldsymbol{\theta}^1, \boldsymbol{\theta}^2, \cdots, \boldsymbol{\theta}^{l_i}\}$, {\mcs} fits a quadratic model with these $3$ points. For the hyper-rectangles sharing $\bs{\theta}^*$, the one containing the minimizer of the quadratic model will be picked as the ``current'' hyper-rectangle for the next dimension.

Such a process continues for $i = 1, 2, \cdots, d_{\theta}$ so as to finish the initialization step.

\subsubsection{Sweep}

After initialization, {\mcs} continues with a sweep process across all the hyper-rectangles according to the following three steps:
\begin{itemize}

\item For all the hyper-rectangles with level $s \in (0, s_{max})$, {\mcs} introduces a pointer $p_s$ indicating the hyper-rectangle belonging to level $s$ with the smallest function evaluation value. If there is no hyper-rectangle of level $s$, then {\mcs} sets $p_s = 0$. 

\item According to the rules to be introduced in the following subsection, for level $s$, if the hyper-rectangle pointed by $p_s$ is not divided, {\mcs} will increase its level by $1$ and update $b_{s+1}$ if necessary. If the hyper-rectangle pointed by $b_s$ is divided, {\mcs} will mark it as divided and insert its children to higher levels. {\mcs} will update the list and the pointers if any children hyper-rectangles introduce a smaller function evaluations.

\item Increase $s$ by $1$. If $s = s_{max}$, start a new sweep; else if $s=0$, go to Step $3$; else go to step $2$.

\end{itemize}

The sweep step outlines the general architecture of the {\mcs} algorithm, and next we will introduce the division rules in the following subsection.

\subsubsection{Hyper-rectangle Division}

Given a candidate hyper-rectangle of level $s < s_{max}$ for division, the division rules include the following two categories:
\begin{enumerate}
\item \textit{Division by Rank}: Let $n_i$ denotes the times coordinate $i$ has been divided in the past for the hyper-rectangle, we can denote $n_{min} = \min_{i \in \{1, 2, \cdots, d_{\theta}\}} n_i$ as the minimum division count for all the coordinates in the search space. If
\begin{equation}
s > 2d_{\theta} (n_{min} + 1),
\end{equation}
the hyper-rectangle will always be divided, and the division dimension index will be $i = \arg \min_{i \in \{1, 2, \cdots, d_{\theta}\}} n_i$. In other words, for the hyper-rectangles which have been divided for many times (with a relatively large level $s$), there may exist some coordinates which haven't been divided very often and {\mcs} will choose to divide these coordinates.

For the selected coordinate $i$, (1) if $n_i = 0$ (i.e., the coordinate hasn't beed divided at all), the division rules will be very similar to the initialization division process (involving the selection of $l_i$ sample points as well as the division points at the golden section). Meanwhile, (2) if $n_i > 0$, {\mcs} will select two division points to divide the hyper-rectangle into three parts. In {\mcs}, to store the hyper-rectangle information, besides the base point $\bs{\theta}$, another \textit{opposite point} $\mb{o}$ will be stored as well, which denotes one of the corners of the hyper-rectangle farthest away from $\bs{\theta}$. Therefore, the range to be divided along the $i_{th}$ dimension will be $[\bs{\theta}_i, \mb{o}_i]$ and the selected division points include: (1) $\mb{z}_i = \bs{\theta}_i + \frac{2}{3}(\xi'' - \bs{\theta}_i)$, and (2) the golden section division point. Here, term 
\begin{equation}\label{equ:mcs_xi_term_subint_definition}
\xi'' = subint(\bs{\theta}_i, \mb{o}_i) = 
\begin{cases}
sign(\mb{o}_i), & \mbox{ if } 1000|\bs{\theta}_i| < 1, |\mb{o}_i| > 1000;\\
10sign(\mb{o}_i)|\bs{\theta}_i|, & \mbox{ if } 1000|\bs{\theta}_i| < 1, |\mb{o}_i| > 1000|\bs{\theta}_i|;\\
\mb{o}_i, &\mbox{ otherwise}.
\end{cases}
\end{equation}
Therefore, with these two points, the hyper-rectangle will be divided into $3$ parts with one more function evaluation at $\mb{z}$ (a point with identical coordinates with $\mb{\theta}$ except coordinate $i$ with $\mb{z}_i$), which will also serve as the base point for the second and third sub-hyper-rectangles. The smaller fraction of the golden section split gets level $\min(s+2, s_{max})$ and the other two get level $s+1$.

\item \textit{Division by Expected Gain}: If 
\begin{equation}
s \le 2d_{\theta} (n_{min} + 1),
\end{equation}
the hyper-rectangle may be divided along a coordinate where the maximum function value gain can be achieved according to a local quadratic model obtained by fitting $2 d_{\theta} + 1$ function values. Meanwhile, if the function expected gain is not large enough, {\mcs} will increase its level by $1$ without any division.

If in the history of the current hyper-rectangle, coordinate $i$ has never been divided, i.e., $n_i = 0$, then {\mcs} will divide the hyper-rectangle along the $i_{th}$ dimension according to the initialization step as introduced before. Meanwhile, if $n_i > 0$, {\mcs} will retrieve the closest two points to the base point in the current hyper-rectangle for each coordinate $i \in \{1, 2, \cdots, d_{\theta}\}$. By fitting a quadratic model with these $3$ points, {\mcs} builds the following function
\begin{equation}
e_i(\xi) = \alpha (\xi - \bs{\theta}_i) + \beta (\xi - \bs{\theta}_i)^2.
\end{equation}

Within interval $[\xi', \xi'']$ ($\xi''$ is defined in Equation~\ref{equ:mcs_xi_term_subint_definition}, and $\xi' = \bs{\theta}_i + (\xi'' - \bs{\theta}_i)/10$), {\mcs} will identify the point achieving the minimum as the potential division point, which can be denoted as
\begin{equation}
\mb{z}_i = \arg \min_{\xi \in [\min(\xi', \xi''), \max(\xi', \xi'')]} e_i(\xi).
\end{equation}
{\mcs} will divide the hyper-rectangle along dimension $i$ at $\mb{z}_i$ if the following condition holds:
\begin{equation}
\mc{L}(\bs{\theta}) + \min_{i \in \{1, 2, \cdots, d_{\theta}\}} \mc{L}(\mb{z}) < \mc{L}^*,
\end{equation}
where $\mc{L}^*$ denotes the current best function value (including the function values obtained by local optimization). Otherwise, {\mcs} will increase the current hyper-rectangle level by $1$ without division.

\end{enumerate}

In addition to the three steps mentioned above, {\mcs} also adopts a local search step, which provides powerful tools for the task of optimization of a smooth function when knowledge about the gradient or even the Hessian is known. Meanwhile, when there exists no information about the derivative information, successive line searches can be adopted. In this paper, we will not introduce the local search step adopted in {\mcs}, and the readers may refer to \cite{mcs} for more information.

\section{Summary}

In this paper, we have introduce two categories of derivative-free optimization algorithms, including the Bayesian methods and Lipschitzian approaches. Many of these introduced algorithms can be potentially applied to learn the deep neural network models. This tutorial paper will also be updated accordingly as we observe more new developments on this topic in the near future.

\newpage

\vskip 0.2in
\bibliographystyle{plain}
\bibliography{reference}

\end{document}